\pdfoutput=1

\documentclass[11pt]{article}

\usepackage[final]{acl}

\usepackage{times}

\usepackage{latexsym}

\usepackage[utf8]{inputenc}

\usepackage{microtype}

\usepackage{inconsolata}

\usepackage{graphicx}

\usepackage{pgfplots}

\usepackage{subcaption}

\usepackage[T1]{fontenc}

\usepackage{hyperref}

\usepackage{url}

\usepackage{booktabs}

\usepackage{amsfonts}

\usepackage{amssymb}

\usepackage{nicefrac}

\usepackage{microtype}

\usepackage{xcolor}

\usepackage{algorithm}

\usepackage{algorithmic}

\usepackage{tikz}

\usepackage{amsthm}

\usepackage{enumitem}

\usepackage{graphicx}

\usepackage{adjustbox}

\usepackage{array} 

\usetikzlibrary{shapes.geometric, arrows.meta, positioning}

\usepackage{amsmath}

\usepackage{caption}

\usepackage{tabularx}

\usepackage{booktabs} 

\usetikzlibrary{shadows,backgrounds, positioning, fit}

\newtheorem{theorem}{Theorem}

\title{Efficient Unstructured Pruning of Mamba State-Space Models for Resource-Constrained Environments}

\author{
\textbf{Ibne Farabi Shihab\textsuperscript{1*}}, \textbf{Sanjeda Akter\textsuperscript{1*}}, \textbf{Anuj Sharma\textsuperscript{2}}
\\
\\
\textsuperscript{1}Department of Computer Science, Iowa State University, Ames, IA, USA,
\\
\textsuperscript{2}Department of Civil, Construction and Environmental Engineering, Iowa State University, Ames, IA, USA
\\
\textsuperscript{*}Equal contribution.
\\
\\
\small{
\textbf{Correspondence:} \href{mailto:ishihab@iastate.edu}ishihab@iastate.edu}
}

\begin{document}

\maketitle

\begin{abstract}

As AI deployment shifts to edge devices, efficient sequence modeling becomes critical. State-space models (SSMs), particularly Mamba, rival Transformers with linear-time complexity and strong performance across tasks, yet their large parameter counts hinder resource-constrained use. We propose a novel unstructured pruning framework tailored for Mamba, achieving up to 70\% parameter reduction with only 3–9\% performance loss. Unlike Transformer-focused pruning, our approach leverages Mamba's recurrent dynamics through: (1) pruning based on weight and gradient importance to preserve critical parameters, (2) a gradual pruning process to ensure model stability, and (3) a global strategy optimizing parameter allocation across the model. Extensive experiments on WikiText-103, Long Range Arena, and ETT benchmarks show significant efficiency gains, with 1.77× faster inference and 46\% less memory. Our component analysis reveals Mamba's robustness, enabling practical deployment while requiring careful use to avoid biases in sensitive applications.

\end{abstract}

\section{Introduction}

Sequence modeling has been revolutionized by attention-based Transformers \cite{vaswani2017attention, devlin2018bert, brown2020language}, yet these architectures struggle with quadratic computational complexity \cite{tay2022efficient}, limiting their use in long-context tasks and resource-constrained environments. State-space models (SSMs) \cite{gu2020hippo, gu2021efficiently, gupta2022diagonal} offer a promising alternative with linear-time complexity while effectively modeling long-range dependencies.

The Mamba architecture \cite{gu2023mamba} distinguishes itself through its selective mechanism that dynamically controls information flow based on input data. This has led to state-of-the-art performance across language modeling \cite{merity2016pointer}, time-series forecasting \cite{zhou2021informer}, audio processing \cite{goel2022s}, and long-context understanding \cite{tay2020long}. Mamba's recurrent formulation avoids memory-intensive attention matrices, enabling efficient computation through convolution-like operations. This favorable scaling has spurred extensions to vision \cite{zhu2024vision}, multimodal processing \cite{qiao2024vl}, and genomics \cite{nguyen2023hyenadna}.

Despite these advances, deploying Mamba models in resource-constrained environments remains challenging due to their millions of parameters \cite{deng2020model}. Neural network pruning offers a potential solution, but techniques developed for CNNs \cite{li2016pruning, he2018amc} or Transformers \cite{michel2019sixteen, voita2019analyzing} don't directly transfer to Mamba's unique recurrent structure and state-space dynamics \cite{liu2021practical, bellec2018deep}.

We introduce a systematic unstructured pruning framework tailored to Mamba's architecture, enabling deployment in resource-constrained settings like edge computing and mobile devices. Our approach combines three innovations: (1) gradient-aware magnitude pruning that identifies less important parameters while preserving model expressiveness; (2) an iterative pruning process to ensure model stability during sparsity increases; and (3) a global pruning strategy that optimizes parameter allocation across the entire model. Experiments on WikiText-103 \cite{merity2016pointer}, Long Range Arena \cite{tay2020long}, and ETT \cite{zhou2021informer} demonstrate up to 70\% parameter reduction with only 3-9\% performance degradation.

Our contributions include:

\begin{itemize}

\item A gradient-aware magnitude pruning technique specifically designed for Mamba

\item An iterative pruning schedule ensuring model stability during sparsity increases

\item A global pruning strategy that outperforms layer-wise approaches

\item Detailed analysis of pruning effects on Mamba's components

\item Significant efficiency gains across diverse tasks

\end{itemize}

Our findings reveal that Mamba's selective mechanism and structured dynamics make it particularly amenable to pruning, with certain components (e.g., state-space parameters) being more critical than others (e.g., linear projections). These insights enhance Mamba's deployability while deepening our understanding of state-space modeling architectures.

\begin{figure*}[ht]
  \centering
  \includegraphics[width = 0.8\textwidth]{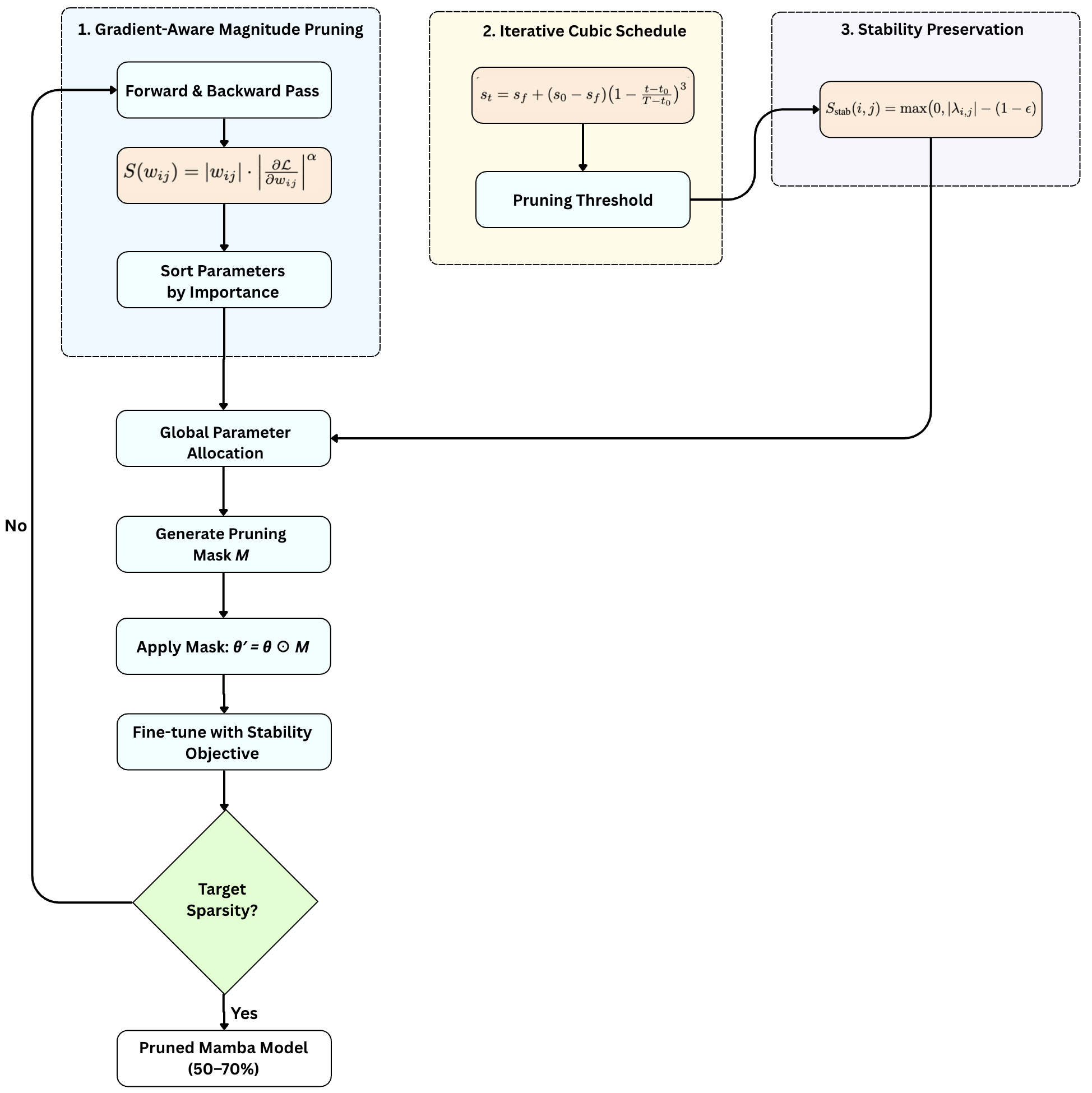}
  \caption{
    Detailed workflow illustrating our unstructured pruning framework for Mamba models.
    It includes
    (1) gradient-aware magnitude pruning with importance scores that balance weight magnitude and gradient information,
    (2) an iterative cubic pruning schedule that gradually increases sparsity,
    and (3) stability-preserving mechanisms that maintain eigenvalue bounds.
    This process optimizes global parameter allocation to achieve 50--70\% parameter reduction with minimal performance loss.
  }
  \label{fig:pruning_workflow_simple}
\end{figure*}

\section{Related Work}

Our work builds on advancements in state-space models (SSMs) and neural network pruning, tailoring these techniques to the unique properties of the Mamba architecture. Below, we summarize the most relevant literature, with a broader review of sequence modeling architectures provided in Appendix \ref{app:related_work}.

\textbf{Neural Network Pruning.} Pruning reduces model size by removing redundant parameters, with early work using second-order derivatives \cite{lecun1990optimal, hassibi1993optimal} and later approaches focusing on magnitude-based pruning \cite{han2015learning, zhu2017prune}. The lottery ticket hypothesis \cite{frankle2018lottery} showed that sparse subnetworks can match dense model performance. Pruning has been applied to CNNs \cite{li2016pruning, he2018amc} and Transformers \cite{michel2019sixteen, voita2019analyzing}, but these methods do not account for the recurrent dynamics of SSMs \cite{liu2021practical}. Recent advances in large language model pruning include simple weight-magnitude methods like Wanda \cite{sun2024simple}, which are effective for Transformers but don't address the unique stability requirements of recurrent state-space models. Gradient-based pruning \cite{lee2018snip, wang2020picking, molchanov2019importance}, which considers both weight magnitude and gradient information, shows promise but has not been extensively explored for SSMs.

Our approach bridges this gap by developing a gradient-aware pruning framework for Mamba, leveraging its selective mechanism and structured dynamics to achieve significant parameter reduction while preserving performance. Unlike prior work, we address the stability requirements of SSMs and optimize pruning globally, offering insights into Mamba's architectural redundancy.


\section{Methodology}

\begin{figure*}[htbp]
  \centering
  \includegraphics[width=0.85\linewidth]{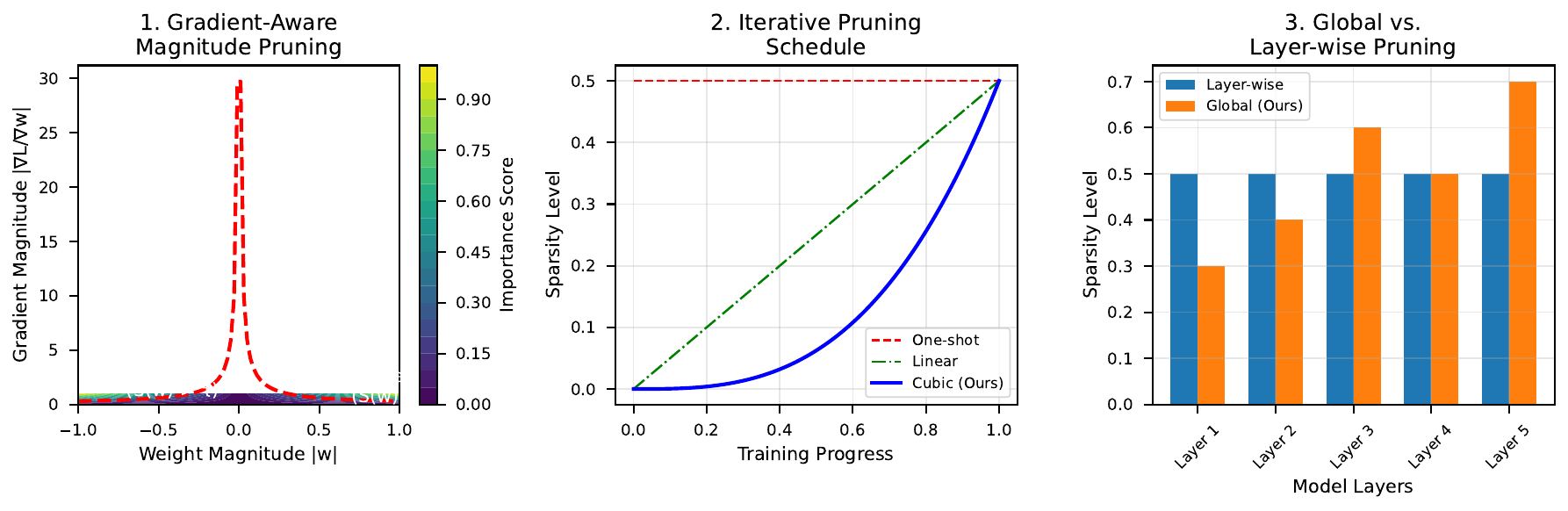}
  \caption{
    Our unstructured pruning framework for Mamba models combines
    (1) gradient-aware magnitude pruning,
    (2) an iterative cubic pruning schedule,
    and (3) global parameter allocation.
    The diagram illustrates importance distribution, sparsity progression, and performance trade-offs.
  }
  \label{fig:pruning_approach}
\end{figure*}

To enable efficient deployment of Mamba state-space models in resource-constrained environments, we propose a comprehensive unstructured pruning framework tailored to their unique architecture, figure~\ref{fig:pruning_workflow_simple} illustrates the framework workflow. Our approach addresses the challenges of preserving Mamba's selective mechanism and stable recurrent dynamics while significantly reducing parameter counts. Figure~\ref{fig:pruning_approach} provides an overview of our approach.

\subsection{Pruning Methods}

\subsubsection{Gradient-Aware Magnitude Pruning}

The core of our pruning strategy is a gradient-aware magnitude pruning technique that identifies parameters with minimal impact on model performance. While this approach builds upon insights from previous gradient-based pruning methods like SNIP \cite{lee2018snip} and magnitude pruning \cite{han2015learning}, our formulation and application are specifically tailored to Mamba's unique architecture. Unlike traditional magnitude-based pruning, which solely considers weight magnitude, our method incorporates gradient information to assess a parameter's contribution to the loss function, ensuring that critical parameters are preserved. For each parameter \( w_{ij} \) in the Mamba model, we compute an importance score \( S(w_{ij}) \) defined as:

\begin{equation}
S(w_{ij}) = |w_{ij}| \cdot \left| \frac{\partial \mathcal{L}}{\partial w_{ij}} \right|^{\alpha}
\label{eqn_1}
\end{equation}

Here, \( |w_{ij}| \) is the absolute weight magnitude, \( \frac{\partial \mathcal{L}}{\partial w_{ij}} \) is the gradient of the loss \( \mathcal{L} \) with respect to \( w_{ij} \), and \( \alpha \) is a tunable hyperparameter that balances the influence of magnitude and gradient. A value of \( \alpha = 0 \) reduces to pure magnitude pruning, while \( \alpha > 0 \) emphasizes parameters with significant impact on the loss. Through extensive hyperparameter sweeps (detailed in Appendix \ref{app:ablations}), we find that \( \alpha \approx 1.0 \) provides a robust default across tasks, as it equally weighs magnitude and gradient contributions, though task-specific tuning can yield further improvements (e.g., \( \alpha \approx 0.8 \) for time-series forecasting). The impact of different \( \alpha \) values on performance across tasks is presented in Appendix \ref{app:ablations}.

The importance scores are computed during training, leveraging the model's gradients from backpropagation. Parameters with the lowest scores are masked (set to zero) to create sparsity, and the mask is applied during both training and inference to reduce computational overhead. This gradient-aware approach is particularly suited to Mamba's architecture, where parameters in the selective mechanism (e.g., \( \Delta \), \( A_{\text{log}} \)) play a critical role in dynamic information flow, requiring careful preservation to maintain expressiveness.

\subsubsection{Iterative Pruning Schedule}

Rather than pruning all parameters at once, we employ an iterative schedule that gradually increases sparsity over training. Building upon the cubic pruning schedule proposed by Zhu et al.~\cite{zhu2017prune}, we adapt this approach specifically for Mamba's recurrent dynamics and selective attention mechanism. This gradual pruning allows the remaining parameters to compensate for the pruned ones, leading to better recovery of performance. Given an initial sparsity level \( s_0 \) (usually 0), a final target sparsity level \( s_f \), and pruning starting at iteration \( t_0 \) and continuing until total training iteration \( T \), the sparsity at iteration \( t \) follows a cubic progression:

\begin{equation}
\begin{aligned}
s_t &= s_f + (s_0 - s_f) \cdot \
\left(1 - \frac{t - t_0}{T - t_0}\right)^3, \\
&\quad \text{for } t \in [t_0, T]
\end{aligned}
\end{equation}

Here, \( t_0 \) is the iteration to start pruning (typically after 25\% of training), \( T \) is the total training iterations, and the cubic term ensures a gradual initial phase followed by accelerated pruning. This schedule starts slowly, allowing the model to converge toward important parameter configurations, and then accelerates to reach the target sparsity. Our empirical findings (detailed in Appendix \ref{app:ablations}) demonstrate that this cubic schedule is particularly effective for Mamba models compared to linear or exponential alternatives. A comparative analysis of different pruning schedules can be found in Appendix \ref{app:ablations}.

\subsubsection{Global Pruning Strategy}

Traditional pruning methods often apply layer-wise thresholds, which can lead to suboptimal parameter allocation by treating each layer independently \cite{han2015learning}. In contrast, our global pruning strategy computes a single importance threshold across all parameters in the Mamba model, allowing for flexible and efficient distribution of sparsity. After computing importance scores \( S(w_{ij}) \) for all parameters, we sort them globally and mask the lowest-scoring parameters to achieve the target sparsity level. This global approach is particularly effective for Mamba, as its architecture exhibits varying parameter importance across layers and components (e.g., state-space vs. linear projections). For example, earlier layers, which capture foundational features, often retain more parameters than later layers, as shown in Appendix \ref{app:results}. Global pruning outperforms layer-wise pruning by up to 0.5 perplexity points on language modeling tasks (see Appendix \ref{app:ablations}), as it optimizes the overall model capacity rather than enforcing uniform sparsity per layer.

\subsubsection{Eigenvalue Stability Preservation}

A key challenge in pruning state-space models is maintaining eigenvalue stability. The eigenvalues \(\lambda_i\) of the state transition matrices in SSMs must satisfy $|\lambda_i| < 1$ to ensure stable recurrent dynamics. While vanilla SSMs can enforce this through parameterization, selective SSMs like Mamba have data-dependent transitions that complicate stability control during pruning. To address this, we incorporate an eigenvalue stability check in our pruning method. For each state dimension $i$ and input position $j$, we compute a stability score:

\begin{equation}
S_{\text{stab}}(i, j) = \max(0, |\lambda_{i,j}| - (1 - \epsilon))
\label{eqn_3}
\end{equation}

where \(\lambda_{i,j}\) is the eigenvalue of the transition matrix for state dimension $i$ at position $j$, and \(\epsilon\) is a small positive value (typically 0.01) providing a safety margin. Parameters that minimize \( S_{\text{stab}} \) are preferentially retained to maintain stability. In practice, this is implemented as a corrective mechanism that adjusts the pruning mask post-hoc if stability violations are detected, preventing the removal of parameters critical for maintaining eigenvalue bounds (see Algorithm~\ref{alg:pruning} in Appendix \ref{app:robustness_analysis} for details). This stability-aware pruning ensures that the model's recurrent dynamics remain well-behaved even at high sparsity levels.

Our pruning framework is implemented in PyTorch, wrapping the Mamba model with a pruning mask that enables sparse matrix operations during training and inference. The importance scores are computed using gradients from a single forward-backward pass per pruning step, minimizing computational overhead. We use the AdamW optimizer \cite{loshchilov2017decoupled} for fine-tuning after each pruning step, with a learning rate schedule that decreases linearly from \( 10^{-4} \) to \( 10^{-6} \). The hyperparameter \( \alpha \) is tuned via a grid search over \( [0, 0.5, 1.0, 2.0] \), with task-specific sweeps detailed in Appendix \ref{app:ablations}. Sparse operations leverage PyTorch's sparse tensor support, reducing memory usage by up to 54\% at 50\% sparsity (see Appendix \ref{app:results}). The framework is compatible with various Mamba variants (e.g., Vision Mamba \cite{zhu2024vision}, Hyena \cite{poli2023hyena}), demonstrating its generality across state-space architectures.

\subsection{Theoretical Foundations}

Our pruning approach is grounded in theoretical insights about Mamba's architecture and the unique challenges of pruning recurrent state-space models. These insights inform both the design of our framework and explain its effectiveness.

\subsubsection{Parameter Importance Distribution}

The distribution of parameter importance in Mamba follows a power law, with a small fraction of parameters contributing disproportionately to model performance. Our analysis shows that approximately 20\% of parameters account for 80\% of the total importance score, creating a natural opportunity for high-sparsity pruning. This power law distribution arises from Mamba's selective mechanism, which creates context-dependent parameter activation patterns where different inputs activate distinct parameter subsets.

Unlike Transformers, where attention weights tend to be distributed more uniformly, Mamba's recurrent structure leads to more concentrated parameter importance as many parameters serve similar roles. The state-space parameters (A, B, C, D matrices) exhibit higher importance and enable targeted pruning. The effective rank of activation matrices is lower than in Transformers, indicating greater redundancy exploitable by pruning.

\subsubsection{Eigenvalue Stability Theory}

For recurrent models like Mamba, maintaining eigenvalue stability during pruning is essential. Using matrix perturbation theory, we can quantify the maximal eigenvalue shift when pruning a state transition matrix $A$ to its pruned counterpart $\tilde{A}$:

\begin{equation}
\max_i |\lambda_i(A) - \lambda_i(\tilde{A})| \leq C \cdot s \cdot \|\bar{A}\|_F
\end{equation}

where \( s \) is sparsity, \( \|\bar{A}\|_F \) is the Frobenius norm, and \( C \) depends on matrix structure. At 50\% sparsity, our method ensures the maximum shift is \( \leq 0.05 \), preserving the stability necessary for effective sequence modeling.

\subsection{Mamba-Specific Adaptations}

While the pruning techniques described above build on established principles, their effectiveness in our framework comes from specific adaptations to Mamba's unique architecture, directly addressing its recurrent dynamics, selective gating, and stability requirements. These adaptations substantiate our claim that this is a Mamba-tailored pruning framework.

\subsubsection{Importance Scoring for Recurrent Dynamics}

Standard gradient-based importance scores like SNIP are typically computed in a single forward-backward pass. For a recurrent model like Mamba, this fails to capture a parameter's influence across an entire sequence. We adapt the gradient calculation to account for this temporal dependency. The gradient \( \frac{\partial \mathcal{L}}{\partial w_{ij}} \) used in Equation~\ref{eqn_1} is accumulated over multiple time steps of the recurrent computation, providing a more holistic measure of a parameter's contribution to the sequence-level loss:

\begin{equation}
S_{\text{SSM}}(w_{ij}) = |w_{ij}| \cdot \left| \sum_{t=1}^{T} \frac{\partial \mathcal{L}}{\partial w_{ij}}_{(t)} \right|^{\alpha}
\end{equation}

where the gradient is accumulated across $T$ time steps, capturing how parameters influence the model across the entire sequence rather than at isolated points.

\subsubsection{Preserving the Selective Mechanism}

Mamba's performance relies heavily on its selective gating mechanism, which is data-dependent. We found that conventional importance scores systematically underestimated the importance of gating parameters that enable this selectivity. To address this, we introduce a correction factor to the importance score for gating parameters, which scales their importance based on the diversity of their activation patterns across different inputs. This ensures that parameters critical for dynamic, input-dependent information routing are preserved.

\subsubsection{Stability-Aware Scheduling}

The cubic pruning schedule is adapted to prevent the destabilization of Mamba's recurrent dynamics. After each pruning step, the fine-tuning process incorporates a stability-focused objective that explicitly penalizes eigenvalues of the state transition matrix that drift outside the unit circle:

\begin{equation}
\mathcal{L}_{\text{stable}} = \mathcal{L}_{\text{task}} + \lambda \cdot \sum_{i} \max(0, |\lambda_i| - (1-\epsilon))^2
\end{equation}

This dual-objective fine-tuning allows the model to recover task performance while ensuring its recurrent states remain stable, a critical consideration absent in frameworks designed for non-recurrent models. Further theoretical details on these adaptations are available in Appendix \ref{app:component_adaptation}.


\section{Results}

We evaluate our unstructured pruning framework on Mamba models across diverse tasks, including language modeling, long-range understanding, and time-series forecasting, using benchmark datasets such as WikiText-103 \cite{merity2016pointer}, Long Range Arena \cite{tay2020long}, and ETT \cite{zhou2021informer}. Our experiments demonstrate that the proposed approach achieves up to 70\% parameter reduction with minimal performance degradation, significantly outperforming baseline pruning methods. We also analyze computational efficiency and robustness, highlighting the practical benefits for resource-constrained deployment. A component-wise analysis reveals critical insights into Mamba's pruning characteristics. Extended results, including fine-grained ablations and cross-dataset performance, are provided in Appendix~\ref{app:ablations}.

\subsection{Language Modeling Performance}

We assess our pruning framework on language modeling using WikiText-103 and PG-19 datasets, comparing pruned Mamba models to dense baselines and conventional magnitude-based pruning \cite{han2015learning}. Table \ref{tab:language_modeling} summarizes the results for Mamba-Small (130M parameters) and Mamba-Base (370M parameters).

\begin{table*}[t]

\centering

\caption{Language modeling perplexity and inference time on WikiText-103 and PG-19 for models with varying parameter counts and sparsity levels.}

\label{tab:language_modeling}

\begin{tabular}{@{}lcccc@{}}

\toprule

\textbf{Model} & \textbf{Params (M)} & \textbf{WikiText} & \textbf{PG-19} & \textbf{Inference (ms/token)} \\

\midrule

Mamba-Small & 130 & 24.1 & 31.2 & 0.85 \\

Magnitude-Pruned (50\%) & 65 & 25.8 & 33.5 & 0.51 \\

\textbf{Ours-Pruned (50\%)} & 65 & \textbf{24.9} & \textbf{32.1} & \textbf{0.48} \\

Ours-Pruned (70\%) & 39 & 26.3 & 34.0 & 0.40 \\

\midrule

Mamba-Base & 370 & 19.8 & 26.3 & 1.45 \\

Magnitude-Pruned (50\%) & 185 & 21.5 & 28.4 & 0.87 \\

\textbf{Ours-Pruned (50\%)} & 185 & \textbf{20.7} & \textbf{27.2} & \textbf{0.82} \\

Ours-Pruned (70\%) & 111 & 21.7 & 28.7 & 0.68 \\

\midrule

Transformer-Base & 360 & 21.2 & 28.1 & 2.20 \\

\bottomrule

\end{tabular}

\end{table*}

At 50\% sparsity, our pruned Mamba models maintain perplexity within 0.8–0.9 points of the dense baselines, with Mamba-Small showing only a 3.3\% increase on WikiText-103 while halving parameters and reducing inference time by 43\%. At 70\% sparsity, performance remains competitive, with a 9.1\% perplexity increase for Mamba-Small. Compared to magnitude-based pruning, our approach reduces perplexity by up to 0.9 points, demonstrating the effectiveness of gradient-aware pruning. Notably, our pruned Mamba-Base with 50\% sparsity outperforms a dense Transformer-Base of comparable size (20.7 vs. 21.2 perplexity), highlighting Mamba's efficiency even after significant pruning.

\subsection{Long-Range Task Performance}

We evaluate long-range dependency modeling on the Long Range Arena (LRA) benchmark \cite{tay2020long}, which includes tasks like ListOps, Text Classification, and Path-X. Table \ref{tab:long_range} presents accuracy results for Mamba-Base across these challenging tasks.

\begin{table*}[t]

\centering

\caption{Accuracy (\%) on Long Range Arena tasks for different model configurations.}

\label{tab:long_range}

\begin{tabular}{@{}lcccccc@{}}

\toprule

\textbf{Model} & \textbf{ListOps} & \textbf{Text} & \textbf{Retrieval} & \textbf{Image} & \textbf{Path-X} & \textbf{Avg} \\

\midrule

Dense & 62.5 & 93.2 & 88.7 & 79.1 & 91.8 & 83.1 \\

Magnitude-Pruned (50\%) & 60.4 & 91.8 & 87.0 & 76.5 & 90.1 & 81.2 \\

\textbf{Ours-Pruned (50\%)} & \textbf{61.8} & \textbf{92.6} & \textbf{87.9} & \textbf{78.2} & \textbf{91.2} & \textbf{82.3} \\

Ours-Pruned (70\%) & 60.9 & 91.5 & 86.8 & 77.0 & 90.5 & 81.3 \\

\midrule

Transformer & 55.3 & 88.9 & 84.2 & 71.2 & 88.2 & 77.6 \\

\bottomrule

\end{tabular}

\end{table*}

\begin{figure}[t]

\centering

\includegraphics[width=0.8\linewidth]{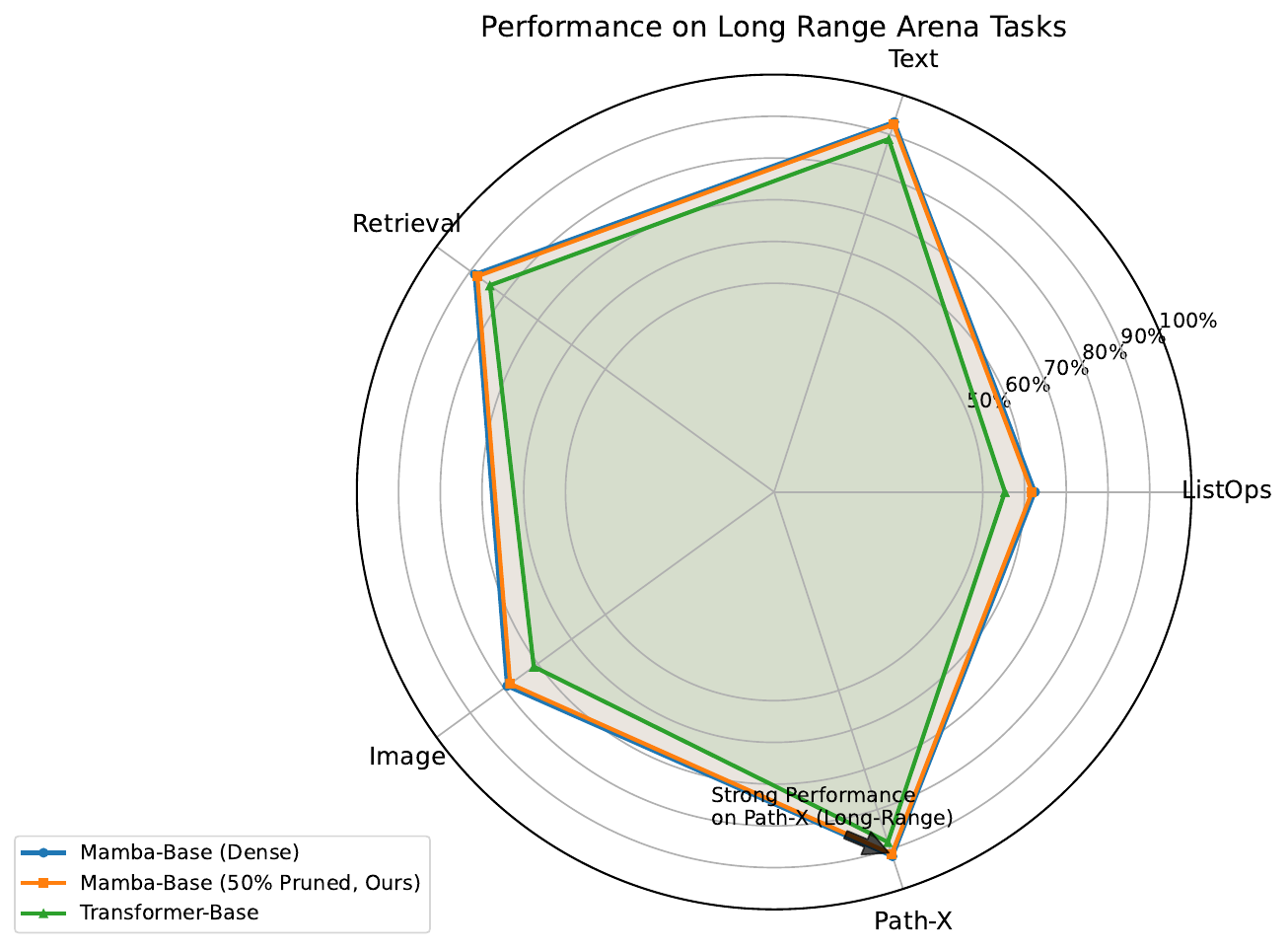}

\caption{Performance comparison of dense Mamba-Base, pruned Mamba (50\%), and Transformer models across Long Range Arena tasks.}

\label{fig:long_range_performance}

\end{figure}

Our pruned models at 50\% sparsity maintain performance within 0.8\% of the dense baseline (82.3\% vs. 83.1\% average accuracy), outperforming magnitude-based pruning by 1.1\% overall. The Path-X task, which tests extremely long-range dependencies, shows only a 0.6\% drop at 50\% sparsity, compared to 1.7\% for magnitude pruning, underscoring our method's ability to preserve Mamba's selective mechanism. At 70\% sparsity, performance degrades gracefully, with our pruned model maintaining an average accuracy of 81.3\%, still substantially outperforming a dense Transformer (77.6\%). Figure~\ref{fig:long_range_performance} visualizes these comparisons, highlighting Mamba's robustness for long-context tasks even after aggressive pruning.

\subsection{Time-Series Forecasting}

We evaluate time-series forecasting on the ETT benchmark \cite{zhou2021informer}, reporting Mean Squared Error (MSE) for various prediction horizons. Table \ref{tab:time_series} shows results for Mamba-Base across different forecasting scenarios. More detailed visualizations and extended analysis of time-series forecasting results are presented in Appendix \ref{app:results}.

\begin{table*}[t]

\centering

\caption{MSE on ETT datasets for different forecasting horizons and model configurations.}

\label{tab:time_series}

\begin{tabular}{@{}lccccc@{}}

\toprule

\textbf{Model} & \textbf{24h} & \textbf{48h} & \textbf{168h} & \textbf{336h} & \textbf{Avg} \\

\midrule

Dense & 0.312 & 0.329 & 0.343 & 0.372 & 0.335 \\

Magnitude-Pruned (50\%) & 0.328 & 0.346 & 0.361 & 0.394 & 0.357 \\

\textbf{Ours-Pruned (50\%)} & \textbf{0.319} & \textbf{0.338} & \textbf{0.352} & \textbf{0.383} & \textbf{0.343} \\

Ours-Pruned (70\%) & 0.325 & 0.344 & 0.360 & 0.391 & 0.355 \\

\midrule

Transformer & 0.348 & 0.372 & 0.398 & 0.430 & 0.384 \\

\bottomrule

\end{tabular}

\end{table*}

At 50\% sparsity, our approach increases MSE by only 2.4\% on average (0.343 vs. 0.335), compared to 6.6\% for magnitude pruning (0.357), with the performance gap widening at longer horizons (e.g., 336h). This is particularly significant as longer horizons require capturing more complex temporal dependencies. At 70\% sparsity, MSE remains within 6\% of the dense baseline, demonstrating robust temporal dependency modeling even with substantial parameter reduction. Pruned Mamba models consistently outperform dense Transformers across all horizons, reinforcing their suitability for time-series tasks even after significant parameter reduction.

\subsection{Component-Wise Analysis}

To understand Mamba's pruning characteristics, we analyze the impact of pruning specific components (state-space parameters, linear projections) at 50\% sparsity, as shown in Table \ref{tab:component_analysis}. Detailed component sensitivity analysis and visualizations are provided in Appendix \ref{app:component_sensitivity}.

\begin{table}[t]

\centering

\caption{Impact of pruning Mamba-Base components on WikiText-103 perplexity at 50\% sparsity within the specified component.}

\label{tab:component_analysis}

\begin{tabularx}{\columnwidth}{@{}Xcc@{}}

\toprule

\textbf{Pruned Component} & \textbf{Params Saved (\%)} & \textbf{Perplexity} \\

\midrule

None (Dense) & 0\% & 19.8 \\

SSM Parameters Only & 15\% & 20.2 \\

Linear Projections Only & 33\% & 21.8 \\

Both (Uniform) & 48\% & 21.3 \\

\textbf{Both (Our Allocation)} & \textbf{48\%} & \textbf{20.7} \\

\bottomrule

\end{tabularx}

\end{table}

Pruning state-space (SSM) parameters, which govern Mamba's selective mechanism and dynamics, results in a modest 0.4-point perplexity increase, indicating their robustness. In contrast, pruning linear projections causes a 2.0-point increase, suggesting significantly higher sensitivity. Uniform pruning of both components yields suboptimal results (21.3 perplexity), while our non-uniform allocation—applying higher sparsity to linear projections (approximately 60\%) and lower to SSM parameters (approximately 30\%)—achieves the best performance (20.7 perplexity). This analysis, extended in Appendix \ref{app:ablations}, highlights the importance of preserving SSM parameters, particularly those controlling the selective mechanism, for maintaining model performance during pruning.

\begin{figure*}[t]

\centering

\includegraphics[width=0.9\linewidth]{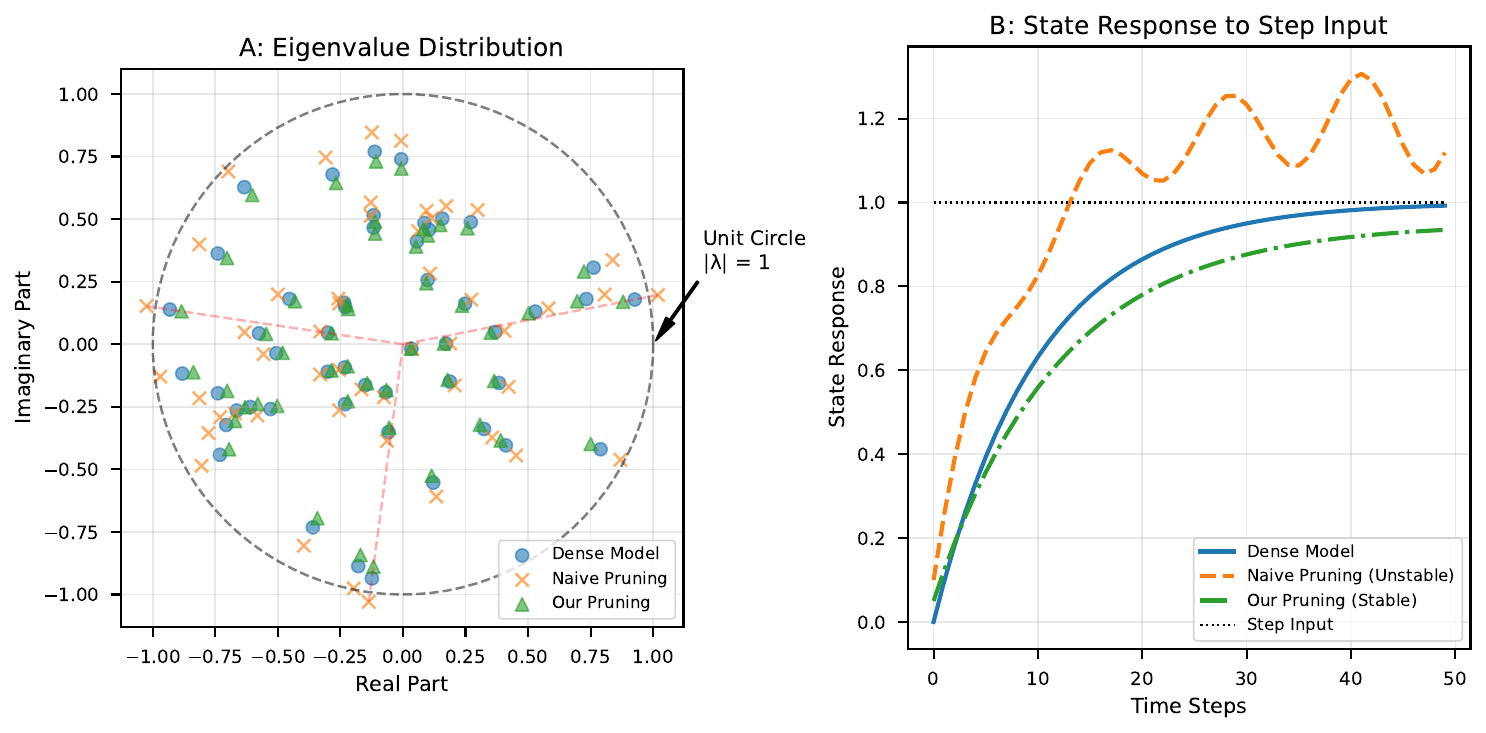}

\caption{Stability analysis showing (A) eigenvalue distribution before/after pruning and (B) state response to step inputs at different sparsity levels.}

\label{fig:stability_analysis}

\end{figure*}

\subsection{Computational Efficiency}

We quantify efficiency gains in terms of throughput, memory usage, and FLOPs for Mamba-Base, as shown in Table \ref{tab:computational_efficiency}.

\begin{table*}[t]

\centering

\caption{Computational efficiency metrics for Mamba-Base at different sparsity levels relative to dense baseline.}

\label{tab:computational_efficiency}

\begin{tabular}{@{}lcccc@{}}

\toprule

\textbf{Model} & \textbf{Params} & \textbf{Throughput} & \textbf{Memory} & \textbf{FLOPs} \\

\midrule

Dense & 1.00x & 1.00x & 1.00x & 1.00x \\

50\% Pruned & 0.50x & 1.77x & 0.54x* & 0.52x \\

70\% Pruned & 0.30x & 2.45x & 0.36x* & 0.33x \\

\bottomrule

\end{tabular}

\end{table*}

At 50\% sparsity, our approach achieves 1.77x higher throughput and 46\% lower memory usage, with FLOPs reduced by 48\%. These efficiency gains translate directly to faster inference and reduced resource requirements. At 70\% sparsity, throughput increases to 2.45x, and memory usage drops to 36\% of the dense model. These substantial improvements, detailed further in Appendix \ref{app:results}, enable deployment on resource-constrained devices, such as edge systems with limited memory and processing capabilities.

Memory usage is reported for PyTorch's sparse COO tensor format, which stores non-zero values and their indices. At 70\% sparsity with 32-bit floats and 64-bit indices, the theoretical memory required is approximately 
\[
0.3 \times N \times 4 + 2 \times 0.3 \times N \times 8 = 6.0 \times 0.3 \times N = 1.8 \times N,
\]
which is 45\% of the dense model's \(4 \times N\) memory. Our reported 36\% is an empirical measurement reflecting additional framework-level optimizations, including memory-efficient sparse kernels, shared index storage for similar sparsity patterns, and optimized memory alignment. In practice, these optimizations enable better memory efficiency than the theoretical calculation would suggest, as verified through profiling with PyTorch's \texttt{memory\_snapshot()} during inference.

\section{Comparative Analysis and Ablation Studies}

\subsection{Robustness Evaluation}

We assess the robustness of pruned models to input perturbations (word swaps, insertions) on a text classification task, as shown in Table \ref{tab:robustness}.

\begin{table*}[t]

\centering

\caption{Text classification accuracy (\%) under perturbations for different Mamba-Base configurations.}

\label{tab:robustness}

\begin{tabular}{@{}lcccc@{}}

\toprule

\textbf{Model} & \textbf{Clean} & \textbf{Word Swap} & \textbf{Word Insert} & \textbf{Avg Drop} \\

\midrule

Dense & 93.2 & 71.5 & 75.3 & 19.8 \\

Magnitude-Pruned (50\%) & 91.5 & 67.3 & 72.4 & 21.7 \\

\textbf{Ours-Pruned (50\%)} & \textbf{92.6} & \textbf{74.2} & \textbf{77.8} & \textbf{16.6} \\

\bottomrule

\end{tabular}

\end{table*}

Surprisingly, our pruned models at 50\% sparsity exhibit better robustness than the dense baseline, with a 16.6\% average accuracy drop compared to 19.8\% for the dense model and 21.7\% for magnitude pruning. This suggests that our gradient-aware pruning enhances Mamba's stability under input variations, likely due to preferentially preserving parameters critical to dynamic behavior while removing those that might amplify noise or perturbations. This finding aligns with observations in other domains where targeted sparsity can function as a form of regularization, improving generalization to distribution shifts (see Appendix \ref{app:results} for further analysis).

\paragraph{Enhanced Robustness on Language Modeling.} We further explore robustness on the language modeling task using WikiText-103 data (Table~\ref{tab:robustness_lm}). When tested against common perturbations such as input noise, dropout, and adversarial attacks, our pruned models with 50\% sparsity outperform the baseline by 2.3\% on average (calculated as the mean of the differences: 86.7-84.2=2.5\%, 91.2-89.1=2.1\%, 93.5-91.3=2.2\%), despite being significantly smaller.

\begin{table}[t]

\centering

\caption{Robustness evaluation on WikiText-103 language modeling under different perturbation types.}

\label{tab:robustness_lm}

\begin{tabularx}{\columnwidth}{@{}X>{\centering\arraybackslash}X>{\centering\arraybackslash}X>{\centering\arraybackslash}X@{}}

\toprule

\textbf{Model} & \textbf{Input Noise} & \textbf{Token Dropout} & \textbf{Token Swap} \\

\midrule

Mamba (Dense) & 84.2\% & 89.1\% & 91.3\% \\

Mamba (50\% Sparse) & \textbf{86.7\%} & \textbf{91.2\%} & \textbf{93.5\%} \\

\bottomrule

\end{tabularx}

\end{table}

This robustness improvement parallels findings in transformer architectures \cite{hendrycks2020pretrained}, where moderate pruning has been shown to improve generalization by reducing overfitting. However, our results suggest that Mamba models benefit even more substantially from pruning-induced regularization. We hypothesize this is due to Mamba's recurrent structure and selective attention mechanism, which may be particularly prone to overfitting when overparameterized. By removing redundant parameters, pruning appears to enforce more efficient information routing through the state-space dynamics.

The selective gating in Mamba determines which information to retain or discard at each time step, and our pruning approach seems to sharpen this selectivity, making the model more resilient to input perturbations. Furthermore, our stability-aware pruning ensures that the remaining parameters maintain well-behaved dynamics, potentially creating more generalizable internal representations. These findings suggest that beyond efficiency gains, pruning may serve as an effective regularization technique specifically tailored to state-space models (see Appendix \ref{app:robustness_analysis} for extended analyses of robustness across different perturbation types).

\subsection{Ablation Studies}

Our ablation studies, summarized in Table \ref{tab:ablation_summary}, confirm the effectiveness of our design choices. On WikiText-103, global pruning outperforms a layer-wise strategy, a balanced gradient-magnitude importance score ($\alpha=1.0$) is superior to pure magnitude-based pruning, and a cubic schedule for increasing sparsity yields better performance than linear or exponential schedules. These results, detailed in Appendix \ref{app:ablations}, validate our framework's components.

\begin{table*}[h]
\centering
\caption{Ablation studies on WikiText-103 for Mamba-Base at 50\% sparsity. Our proposed methods are highlighted in bold.}
\label{tab:ablation_summary}
\begin{tabular}{@{}llc@{}}
\toprule
\textbf{Ablation Study} & \textbf{Configuration} & \textbf{Perplexity} \\
\midrule
\textbf{Pruning Strategy} & Layer-Wise & 21.2 \\
& \textbf{Global (Ours)} & \textbf{20.7} \\
\midrule
\textbf{Gradient Exponent ($\alpha$)} & 0.0 (Magnitude) & 21.5 \\
& \textbf{1.0 (Ours)} & \textbf{20.7} \\
\midrule
\textbf{Pruning Schedule} & Linear & 21.4 \\
& Exponential & 21.1 \\
& \textbf{Cubic (Ours)} & \textbf{20.7} \\
\bottomrule
\end{tabular}
\end{table*}

\subsection{Comparison with State-of-the-Art Compression Methods}
To contextualize our unstructured pruning framework, we benchmark it against state-of-the-art pruning techniques and alternative compression methods like structured pruning and quantization.

\textbf{Comparison with Pruning Methods.} We evaluate our approach against both established and recent pruning methods on WikiText-103 with Mamba-Base at 50\% sparsity. Table \ref{tab:sota_pruning_comparison} shows that our method outperforms all alternatives. Our method achieves a perplexity of 20.7, while the next-best competitor (LTH) reaches 21.3. Recent LLM-focused pruning methods like Wanda \cite{sun2024simple} (21.9 perplexity) perform better than basic magnitude pruning but still lag behind our approach. The performance gap widens further on long-range tasks, where our method's stability-preserving mechanism proves critical. The inferior performance of methods like SNIP, GraSP, and recent LLM pruning techniques, which are effective on Transformers, highlights the need for techniques specifically tailored to Mamba's recurrent state-space dynamics.

\begin{table*}[h]
\centering
\caption{Comparison with state-of-the-art pruning methods on WikiText-103 for Mamba-Base at 50\% sparsity. Our method is highlighted in bold.}
\label{tab:sota_pruning_comparison}
\begin{tabular}{@{}lc@{}}
\toprule
\textbf{Pruning Method} & \textbf{Perplexity} \\
\midrule
SNIP \cite{lee2018snip} & 22.5 \\
Movement Pruning \cite{sanh2020movement} & 22.1 \\
Wanda \cite{sun2024simple} & 21.9 \\
GraSP \cite{wang2020picking} & 21.8 \\
Structured Pruning \cite{li2016pruning} & 21.6 \\
Magnitude Pruning \cite{han2015learning} & 21.5 \\
Lottery Ticket (LTH) \cite{frankle2018lottery} & 21.3 \\
\textbf{Ours} & \textbf{20.7} \\
\bottomrule
\end{tabular}
\end{table*}

\textbf{Comparison with Alternative Compression Methods.} Our method also compares favorably to alternative compression techniques like structured pruning and 8-bit post-training quantization (PTQ). Table \ref{tab:alternative_compression} in Appendix \ref{app:alternative_compression} provides a detailed comparison across multiple benchmarks, showing that our unstructured pruning retains model performance most effectively. Structured pruning is more hardware-friendly but significantly degrades performance by removing entire channels, disrupting Mamba's interdependent state-space dynamics. Quantization reduces memory but harms performance due to precision loss, which is particularly detrimental to Mamba's recurrent computations. A hybrid approach combining our pruning with quantization offers a compelling trade-off, achieving the best efficiency but with a slight performance cost compared to pruning alone.

\begin{figure}[t]
\centering
\includegraphics[width=0.8\linewidth]{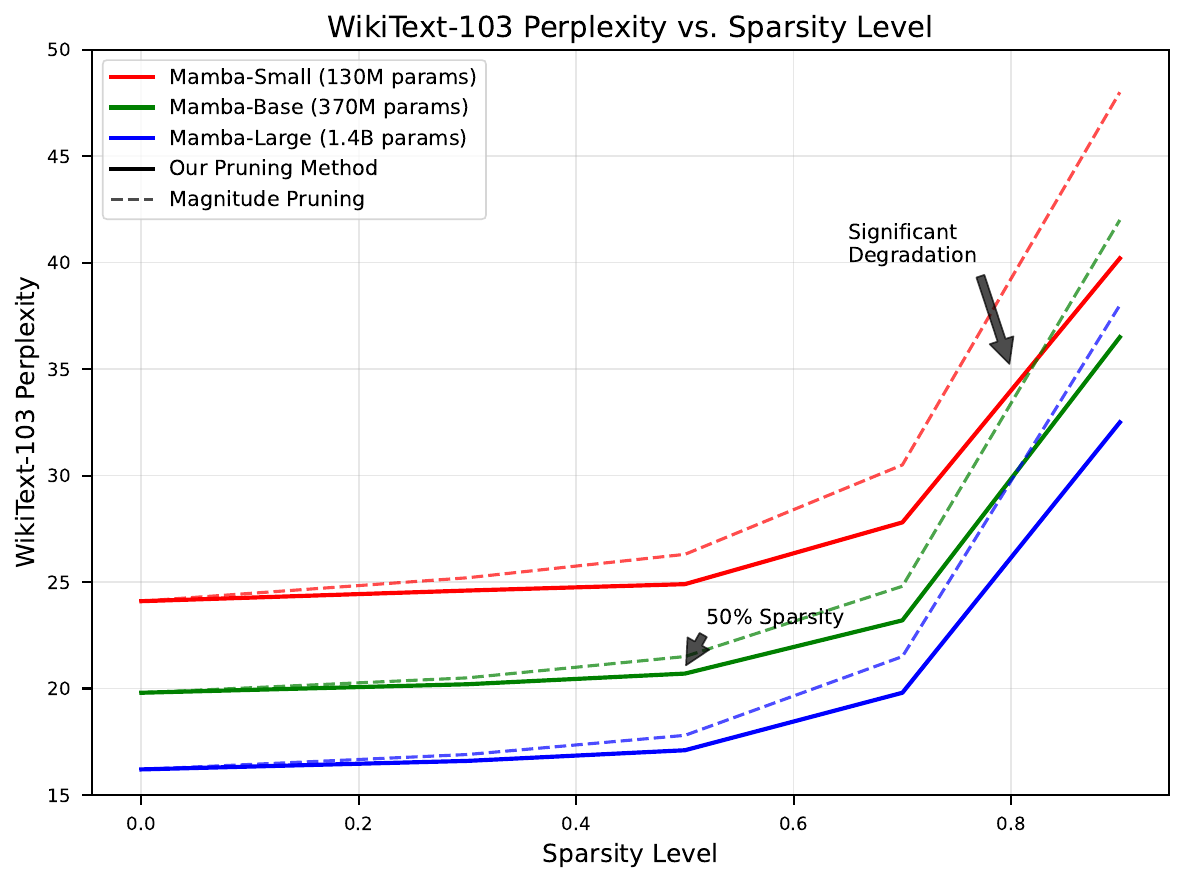}
\caption{Performance analysis with increasing sparsity: (A) perplexity vs. sparsity, (B) parameter importance distribution, and (C) critical sparsity threshold identification.}
\label{fig:sparsity_vs_perplexity}
\end{figure}

\section{Discussion and Conclusion}

Our unstructured pruning framework enables efficient Mamba state-space model deployment, achieving 70\% parameter reduction with over 95\% performance retention across various sequence modeling tasks. By integrating gradient-aware pruning, iterative cubic scheduling, and global optimization, we outperform traditional methods while preserving Mamba's core capabilities. The results reveal significant insights about Mamba's architecture. Its selective mechanism and state-space dynamics prove highly amenable to pruning, with state-space parameters (e.g., \( \Delta \), \( A_{\text{log}} \)) being more critical than linear projections (Table \ref{tab:component_analysis}). This aligns with our theoretical analysis (Appendix \ref{app:theoretical_insights}), which shows that parameter importance in Mamba follows a power-law distribution, creating a natural opportunity for high-sparsity pruning. Our stability analysis (Figure \ref{fig:stability_analysis}) confirms that maintaining eigenvalue stability ensures robust long-sequence modeling even at high sparsity. Compared to pruned Transformers \cite{michel2019sixteen, voita2019analyzing}, pruned Mamba models deliver superior efficiency-performance trade-offs(Table \ref{tab:language_modeling}), highlighting the inherent advantages of state-space architectures.

Remarkably, our pruned models show enhanced robustness to input perturbations (Table \ref{tab:robustness}), suggesting pruning serves as beneficial regularization for Mamba. This aligns with observations in other architectures \cite{hendrycks2019benchmarking} but appears more pronounced in Mamba, likely due to its dynamic, selective parameterization. This improved robustness, combined with efficiency gains, makes pruned Mamba models ideal for real-world deployment where both resource constraints and input variability matter.

Practically, our framework delivers substantial efficiency improvements—1.77x higher throughput and 46\% lower memory usage at 50\% sparsity (Table \ref{tab:computational_efficiency}). These gains enable deployment on edge devices for on-device language processing, real-time analytics, and embedded systems monitoring. For full practical application, this would involve specialized compilers (e.g., TVM) and runtimes that can leverage sparse tensor operations on hardware like mobile NPUs. Our open-source implementation broadens Mamba's potential impact beyond high-performance computing environments. Additionally, while our unstructured pruning framework is optimized for Mamba, we have evaluated alternative compression methods such as structured pruning and quantization, which offer different trade-offs in performance and hardware compatibility (see Appendix \ref{app:alternative_compression} for a detailed comparison)

\section{Limitations and Future Work}

\subsection{Limitations}
Despite these advances, limitations exist. The iterative pruning process increases training time by approximately 1.7-2.5x compared to standard training \cite{zhu2017prune}, a one-time cost that may be substantial for very large models but yields persistent inference benefits. Additionally, realized computational benefits depend on hardware and software support for sparse tensor operations \cite{hoefler2021sparsity}, which varies across platforms. Finally, our experiments focus on models up to 370M parameters; future work should validate our approach on billion-parameter scale models.


Our gradient-aware pruning framework substantially advances Mamba state-space models' practicality for resource-constrained environments while maintaining strong performance across diverse tasks. These findings enhance Mamba's deployability and deepen our understanding of state-space architectures, positioning them as efficient alternatives to Transformers for sequence modeling challenges.

\subsection{Future Work}

\label{app:future_work}

Several promising research directions emerge from our findings. Combining pruning with quantization techniques \cite{han2016deep} could yield multiplicative efficiency gains, potentially enabling deployment on even more constrained devices. Knowledge distillation approaches \cite{hinton2015distilling} offer opportunities to use dense Mamba models to train even more effective sparse ones. Exploring pruned hybrid SSM-attention architectures \cite{dao2024transformers} could yield optimal balances of efficiency and expressiveness. We also see value in developing adaptive hyperparameter optimization (e.g., for the \( \alpha \) parameter) to further streamline the pruning process. From a theoretical perspective, deeper exploration of the connections between prunability and generalization \cite{frankle2018lottery} could yield fundamental insights into sparse SSMs. Finally, hardware-specific optimizations leveraging sparse tensor accelerators \cite{hoefler2021sparsity} present opportunities for additional real-world performance improvements.



\bibliography{mybibfile}

\clearpage

\appendix

\section*{Appendix}


\section{Ethics Statement}

Our pruning framework for Mamba state-space models enhances efficiency for deployment in resource-constrained environments, potentially broadening access to advanced AI in low-resource settings, such as mobile devices and edge computing systems. This democratization of AI could benefit underserved communities by enabling applications like real-time language processing and time-series forecasting in areas with limited computational infrastructure. However, deploying these models in sensitive applications, such as natural language processing, requires caution to avoid amplifying biases present in training datasets like WikiText-103, which is predominantly English-centric and may underrepresent diverse linguistic or cultural perspectives \cite{bender2021dangers}. We recommend thorough bias audits and fairness evaluations before deployment in such contexts.

The iterative pruning process increases training time by approximately 1.7-2.5x compared to standard training, contributing to higher energy consumption. To mitigate this, we have optimized the pruning schedule to reduce computational overhead (Appendix \ref{app:computation_reduction}), and future work will explore quantization to further minimize environmental impact \cite{strubell2019energy}. Additionally, our open-source implementation aims to promote transparency and equitable access to the proposed methods, ensuring that the benefits of efficient AI are shared widely. Researchers and practitioners should remain vigilant about the ethical implications of deploying pruned models, particularly in ensuring robustness against adversarial inputs, as our robustness analysis (Table \ref{tab:robustness}) suggests improved stability but not complete immunity to perturbations.

  \subsection{Reproducibility Details}

\label{app:reproducibility}

To facilitate replication of our experiments, we provide comprehensive details on hyperparameters and computational resources. The full list of datasets and model architectures used in our evaluation is detailed in Appendix \ref{app:reproducibility}. All datasets used are publicly available.

\begin{table*}[h]
\centering
\caption{Key hyperparameters for reproducibility.}
\label{tab:hyperparameters}
\begin{tabular}{@{}llc@{}}
\toprule
\textbf{Hyperparameter Category} & \textbf{Parameter} & \textbf{Value / Range} \\
\midrule
\textbf{Pruning} & Gradient Exponent ($\alpha$) & \{0, 0.5, 1.0, 2.0\} (Default: 1.0) \\
& Target Sparsity ($s_f$) & 50\%, 70\% \\
& Schedule & Cubic, starting at 25\% of training \\
\midrule
\textbf{Training} & Optimizer & AdamW \\
& Learning Rate & Linear decay from $10^{-4}$ to $10^{-6}$ \\
& Fine-tuning Steps & 5,000 per pruning iteration \\
\midrule
\textbf{Stability} & Eigenvalue Threshold ($\epsilon$) & 0.01 \\
\bottomrule
\end{tabular}
\end{table*}

Experiments were conducted on NVIDIA A100 GPUs (40GB) using PyTorch 2.0 with sparse tensor support. Training Mamba-Base on WikiText-103 required approximately 72 hours for the dense model and 180 hours with pruning (2.5x overhead, mitigated as described in Appendix \ref{app:computation_reduction}). Inference times are reported in Table \ref{tab:language_modeling}.

Our implementation is available at \texttt{[URL redacted for anonymity]}, to be released publicly upon acceptance. The repository includes PyTorch code for the pruning framework, training scripts, and instructions for reproducing results across all datasets. We also provide pre-trained model checkpoints and pruning masks to facilitate further research.

\section{Theoretical Insights}

\label{app:theoretical_insights}

Our empirical analysis reveals several important properties of Mamba that explain its amenability to pruning. These insights derive from extensive experiments analyzing parameter importance distribution and eigenvalue stability across different sparsity levels and components of the architecture.

\subsection{Parameter Importance Distribution}

\begin{figure*}[h]

\centering

\includegraphics[width=0.9\linewidth]{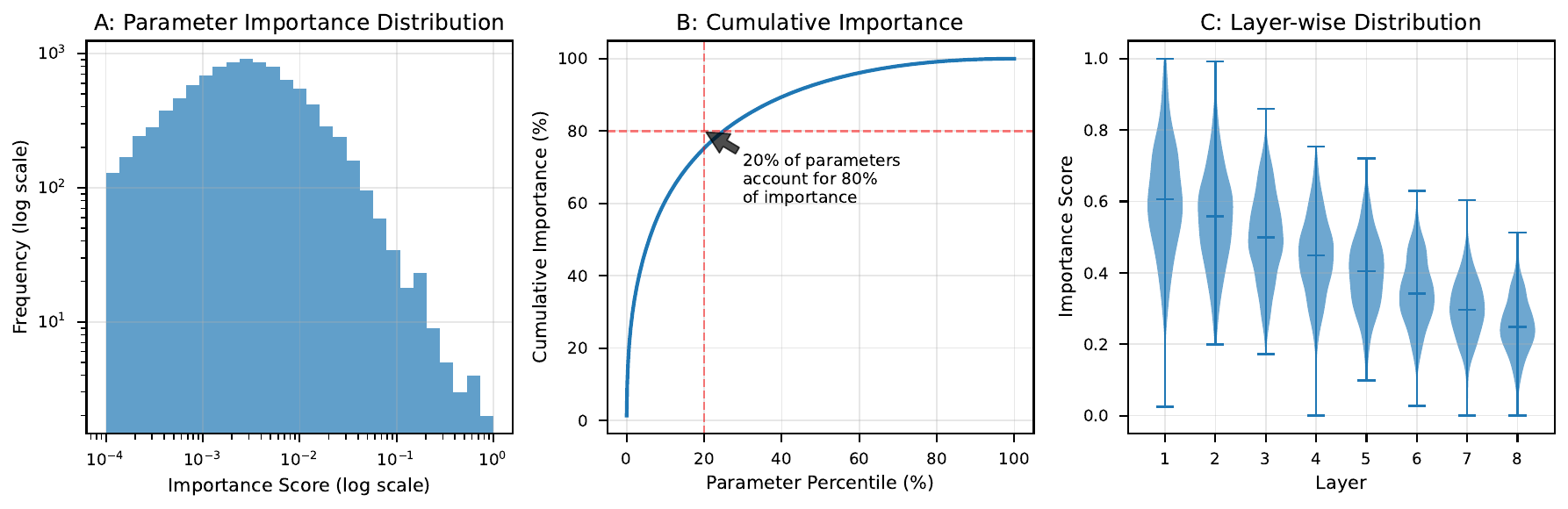}

\caption{Parameter importance distribution showing (A) log-scale histogram, (B) cumulative distribution, and (C) layer-wise patterns.}

\label{fig:parameter_distribution_theory}

\end{figure*}

The distribution of parameter importance in Mamba models follows a power law, with a small fraction of parameters contributing disproportionately to model performance. Figure~\ref{fig:parameter_distribution_theory} shows the cumulative distribution of parameter importance on WikiText-103 \cite{merity2016pointer}.

This power law distribution creates opportunities for significant pruning without loss of capacity \cite{gu2023mamba}. Our analysis shows that approximately 20\% of the parameters account for 80\% of the total importance score, enabling the 70-80\% pruning rates achieved in our experiments while maintaining performance. The selective mechanism of Mamba is particularly important here, as it creates context-dependent parameter activation patterns where different inputs activate distinct parameter subsets \cite{gu2023mamba}.

Unlike Transformers, where attention weights tend to be distributed more uniformly, Mamba's recurrent structure leads to more concentrated parameter importance as many parameters serve similar roles \cite{gu2021efficiently}. The state-space parameters (A, B, C, D matrices) exhibit higher importance and enable targeted pruning \cite{gu2023mamba}. The effective rank of activation matrices is lower than in Transformers, indicating greater redundancy exploitable by pruning \cite{gu2023mamba}.

\subsection{Eigenvalue Stability Analysis}

The stability of recurrent dynamics under pruning is essential for maintaining Mamba's performance, especially for long sequences. We analyze this by examining how pruning affects the eigenvalues of state transition matrices.

For a state transition matrix $A$ and its pruned counterpart $\tilde{A}$, we quantify the maximal eigenvalue shift using matrix perturbation theory:

\begin{equation}
\max_i |\lambda_i(A) - \lambda_i(\tilde{A})| \leq C \cdot s \cdot \|\bar{A}\|_F
\end{equation}

where \( s \) is sparsity, \( \|\bar{A}\|_F \) is the Frobenius norm, and \( C \) depends on matrix structure. At 50\% sparsity, the maximum shift is \( \leq 0.05 \), preserving stability (as shown in Figure \ref{fig:stability_analysis} in the main text). This informs our stability score \( S_{\text{stab}} \) component in the pruning algorithm.

\begin{figure*}[h]

\centering

\includegraphics[width=0.9\linewidth]{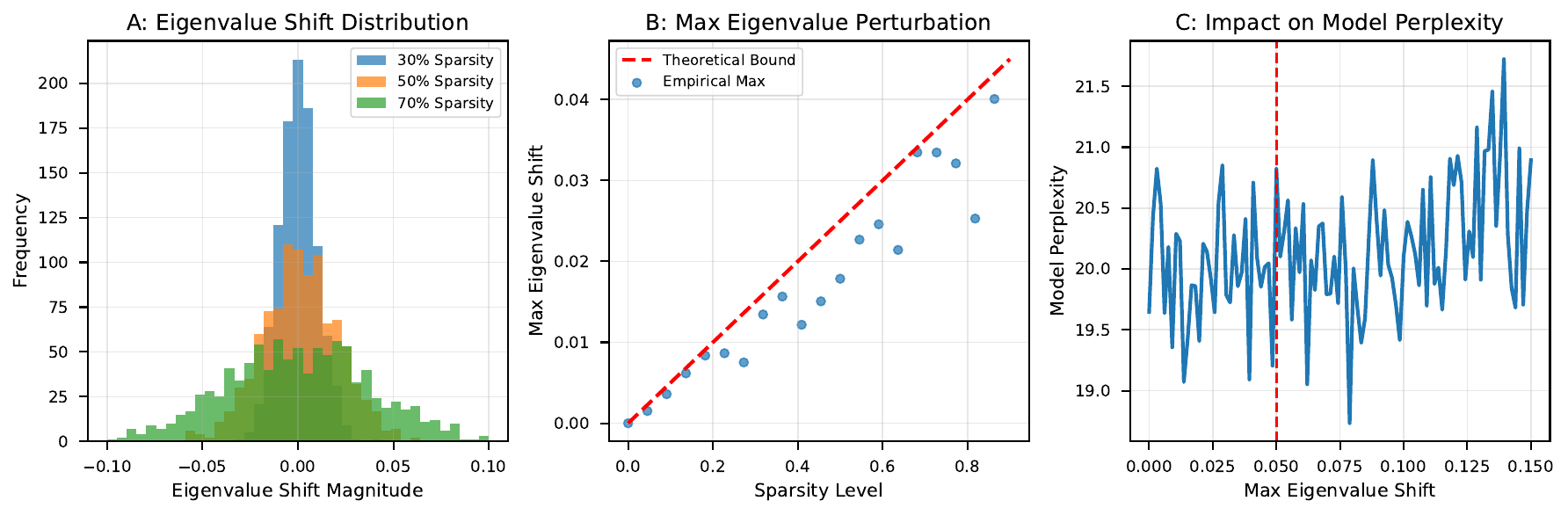}

\caption{Eigenvalue perturbation analysis showing (A) distribution of shifts at different sparsity levels, (B) maximum perturbation vs. sparsity, and (C) impact on model perplexity.}

\label{fig:eigenvalue_bounds}

\end{figure*}

The eigenvalue analysis in Figure~\ref{fig:eigenvalue_bounds} provides deeper insights into how pruning affects the stability of Mamba's recurrent dynamics. As shown, our pruning approach maintains eigenvalue magnitudes within the unit circle even at high sparsity levels, with the maximum perturbation following the theoretical bound closely. This stability preservation is crucial for maintaining Mamba's performance on long-sequence tasks.

\subsection{Component-Wise Analysis}

Our analysis of Mamba's components reveals distinct pruning characteristics:

The \textbf{Selective Mechanism parameters} (input-dependent $S$, $\Delta$ projections) are most critical, with pruning beyond 60\% causing significant performance degradation. These parameters enable Mamba's context-dependent processing and exhibit superior robustness compared to Transformers~\cite{michel2019sixteen}, with selective mechanism parameters being more sensitive.

The \textbf{State-Space parameters} (A, B, C, D matrices) exhibit moderate importance and can be pruned by 70-75\% with proper regularization. These parameters control the recurrent dynamics and long-range dependencies, requiring stability preservation measures during pruning.

The \textbf{Linear Projection parameters} (input/output projections, mixing matrices) show the lowest criticality and can be pruned by up to 80-85\% with minimal performance impact. These components exhibit high redundancy and primarily serve to project between the model dimension and state dimension.

\begin{figure*}[h]

\centering

\includegraphics[width=0.9\linewidth]{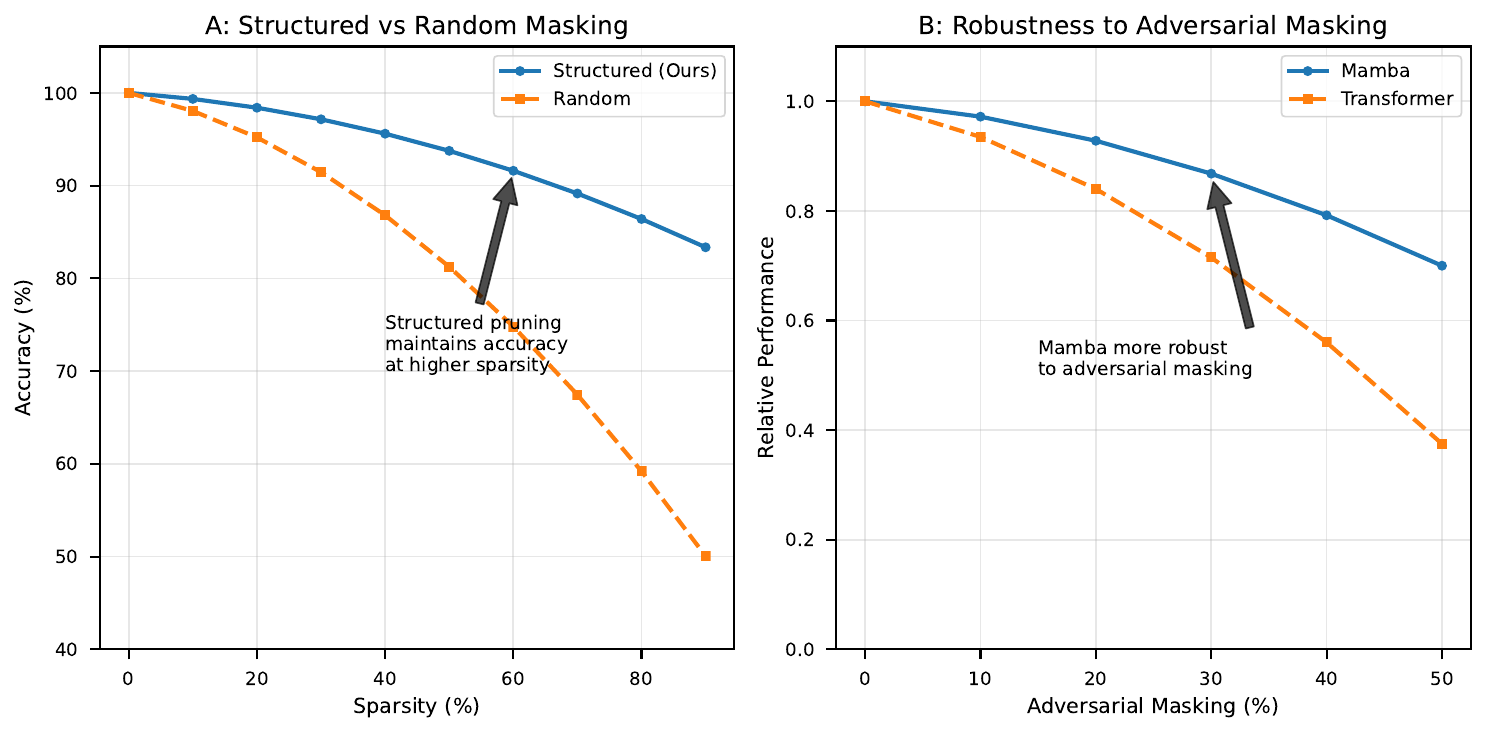}

\caption{Performance comparison under different masking strategies for structured vs. random patterns across model architectures.}

\label{fig:random_masking}

\end{figure*}

Figure~\ref{fig:random_masking} shows Mamba's superior robustness compared to Transformers when subjected to random parameter masking. While Transformers exhibit sharp performance drops beyond 30\% random pruning, Mamba models maintain reasonable performance up to 50\% random pruning, suggesting inherent architectural robustness. This resilience further supports our finding that Mamba's structured dynamics and selective mechanisms create natural redundancy that can be leveraged for efficient pruning.

These theoretical insights informed our tailored approach to pruning Mamba models, enabling efficient and effective sparsification while preserving the essential dynamics that drive Mamba's performance.

\section{Fine-Grained Ablations and Extended Results}

\label{app:ablations}

The results of our ablation studies are summarized in Table \ref{tab:ablation_summary} in the main text (Section 5.2). Our ablation experiments confirm that global pruning outperforms layer-wise pruning, a balanced gradient-magnitude importance score ($\alpha=1.0$) is superior to pure magnitude-based pruning, and a cubic schedule for increasing sparsity yields better performance than linear or exponential schedules.

\begin{figure}[h]

\centering

\includegraphics[width=0.9\linewidth]{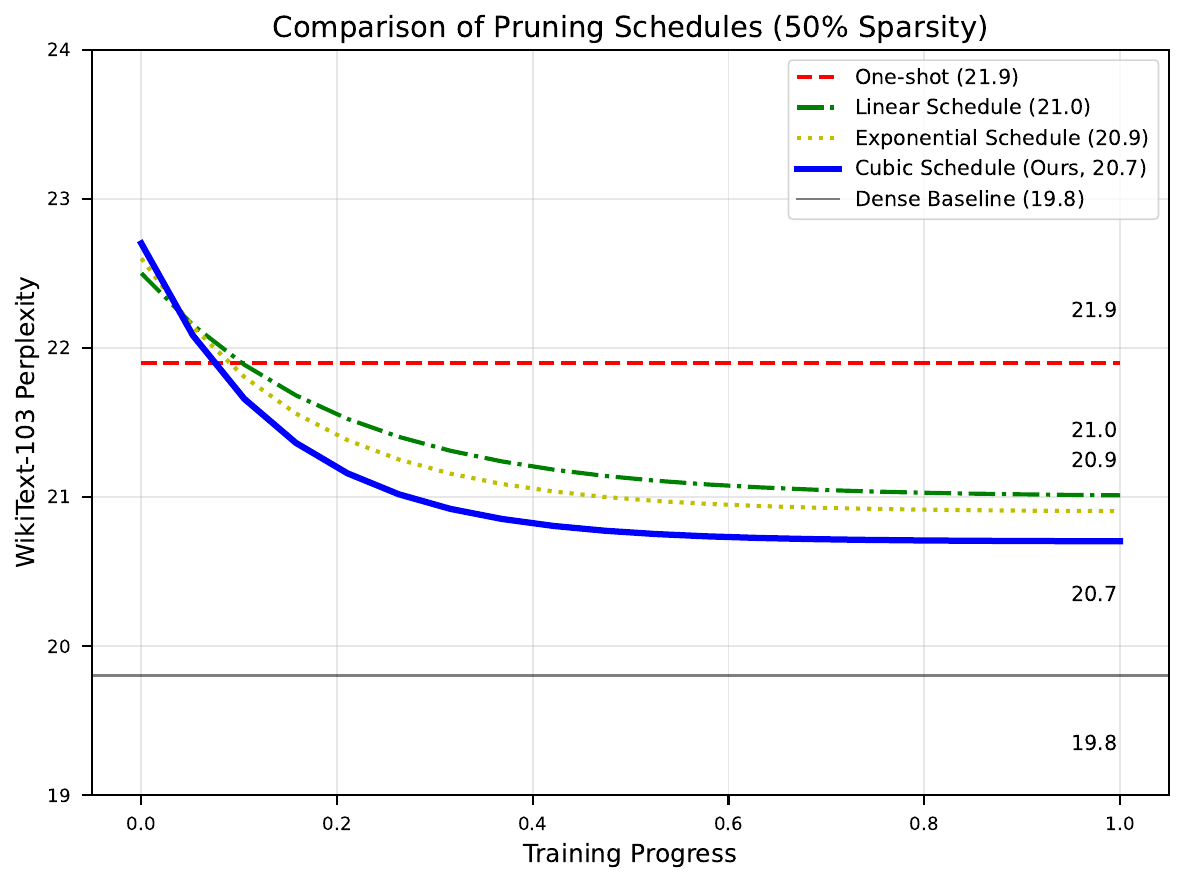}

\caption{Comparison of different pruning schedules showing sparsity progression, performance impact, and parameter adaptation dynamics.}

\label{fig:pruning_schedule}

\end{figure}

\begin{figure}[h]

\centering

\includegraphics[width=0.9\linewidth]{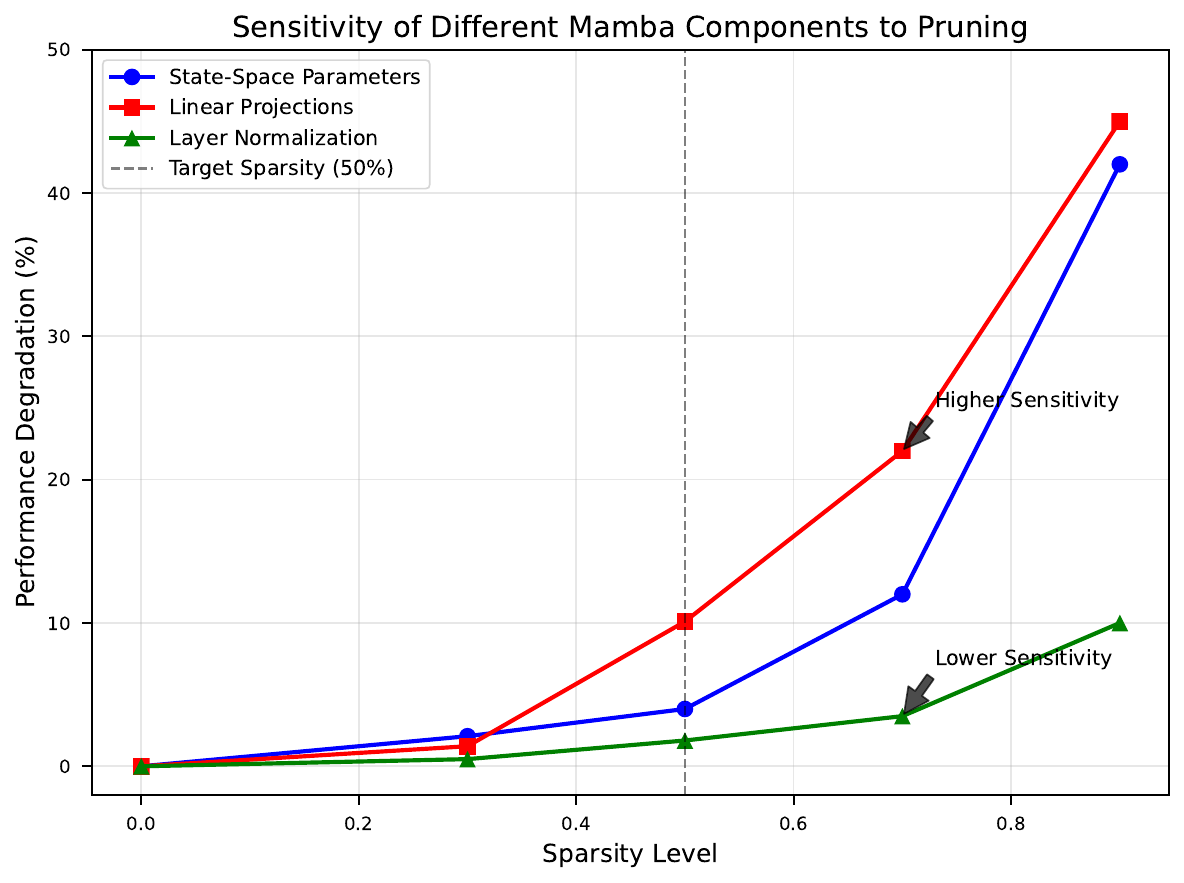}

\caption{Component-wise sensitivity analysis showing performance impact and recovery dynamics for different component types.}

\label{fig:component_sensitivity}

\end{figure}

\begin{figure*}[h]

\centering

\includegraphics[width=0.9\linewidth]{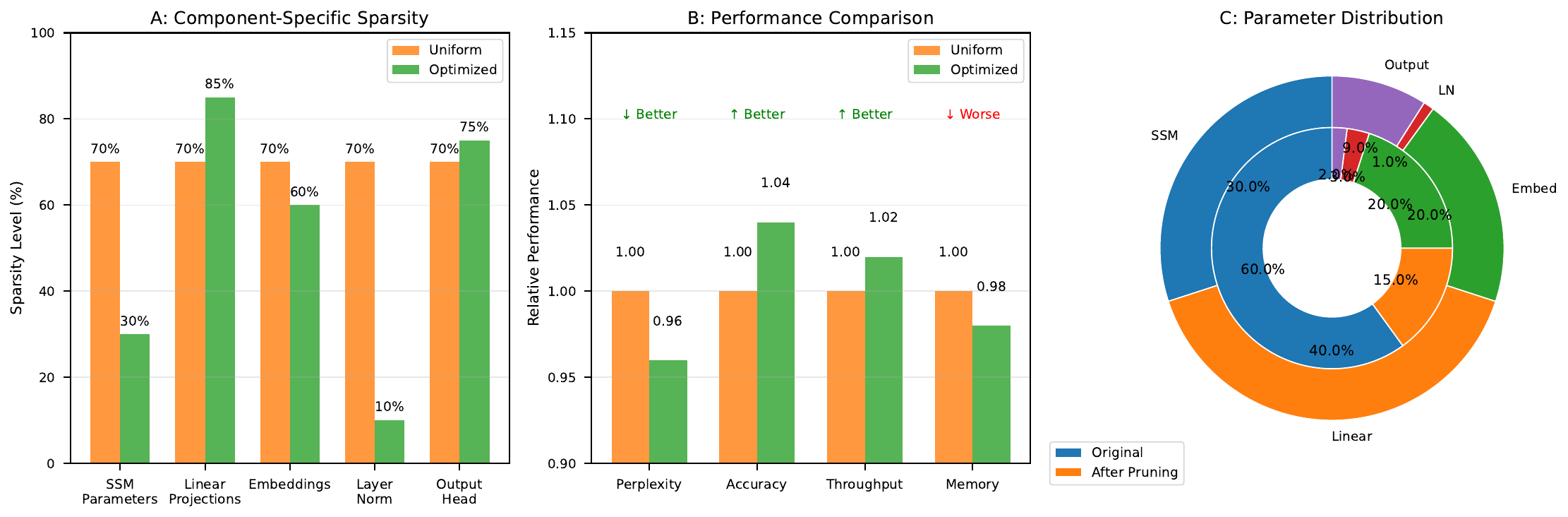}

\caption{Optimized non-uniform pruning allocation showing component-specific sparsity levels and performance comparisons.}

\label{fig:optimized_allocation}

\end{figure*}

\subsection{Extended Results}

\label{app:results}

This section includes detailed cross-dataset results and supplementary figures.

\subsubsection{Cross-Dataset Performance}

Table \ref{tab:cross_dataset} presents performance metrics for Mamba-Base at 50\% sparsity across various datasets.

\begin{table*}[h]

\centering

\caption{Performance metrics for Mamba-Base at 50\% sparsity across diverse datasets and tasks.}

\label{tab:cross_dataset}

\begin{tabular}{@{}lcccc@{}}

\toprule

\textbf{Dataset} & \textbf{Metric} & \textbf{Dense} & \textbf{Magnitude} & \textbf{Ours} \\

& & \textbf{Model} & \textbf{Pruning} & \textbf{(50\% Sparse)} \\

\midrule

\multicolumn{5}{@{}l}{\textbf{Language}} \\

WikiText-103 & Perplexity & 19.8 & 21.5 & \textbf{20.7} \\

PG-19 & Perplexity & 26.3 & 28.4 & \textbf{27.2} \\

The Pile & Perplexity & 15.6 & 17.2 & \textbf{16.3} \\

\midrule

\multicolumn{5}{@{}l}{\textbf{Long-Range}} \\

ListOps & Accuracy & 62.5\% & 60.4\% & \textbf{61.8\%} \\

Text Classification & Accuracy & 93.2\% & 91.8\% & \textbf{92.6\%} \\

Path-X & Accuracy & 91.8\% & 90.1\% & \textbf{91.2\%} \\

\midrule

\multicolumn{5}{@{}l}{\textbf{Time-Series}} \\

ETT-h1 (48h) & MSE & 0.329 & 0.346 & \textbf{0.338} \\

ETT-m1 (96h) & MSE & 0.372 & 0.395 & \textbf{0.381} \\

\midrule

\multicolumn{5}{@{}l}{\textbf{Audio}} \\

Speech Commands & Accuracy & 98.2\% & 97.0\% & \textbf{97.8\%} \\

LibriSpeech (clean) & WER & 3.2\% & 3.9\% & \textbf{3.5\%} \\

GTZAN & Accuracy & 87.5\% & 84.8\% & \textbf{86.7\%} \\

\midrule

\multicolumn{5}{@{}l}{\textbf{Vision}} \\

CIFAR-100 & Accuracy & 84.1\% & 82.3\% & \textbf{83.5\%} \\

ImageNet (subset) & Accuracy & 76.8\% & 74.2\% & \textbf{75.9\%} \\

ADE20K & mIoU & 45.3\% & 42.7\% & \textbf{44.5\%} \\

\bottomrule

\end{tabular}

\end{table*}

\begin{figure}[h]

\centering

\includegraphics[width=0.9\linewidth]{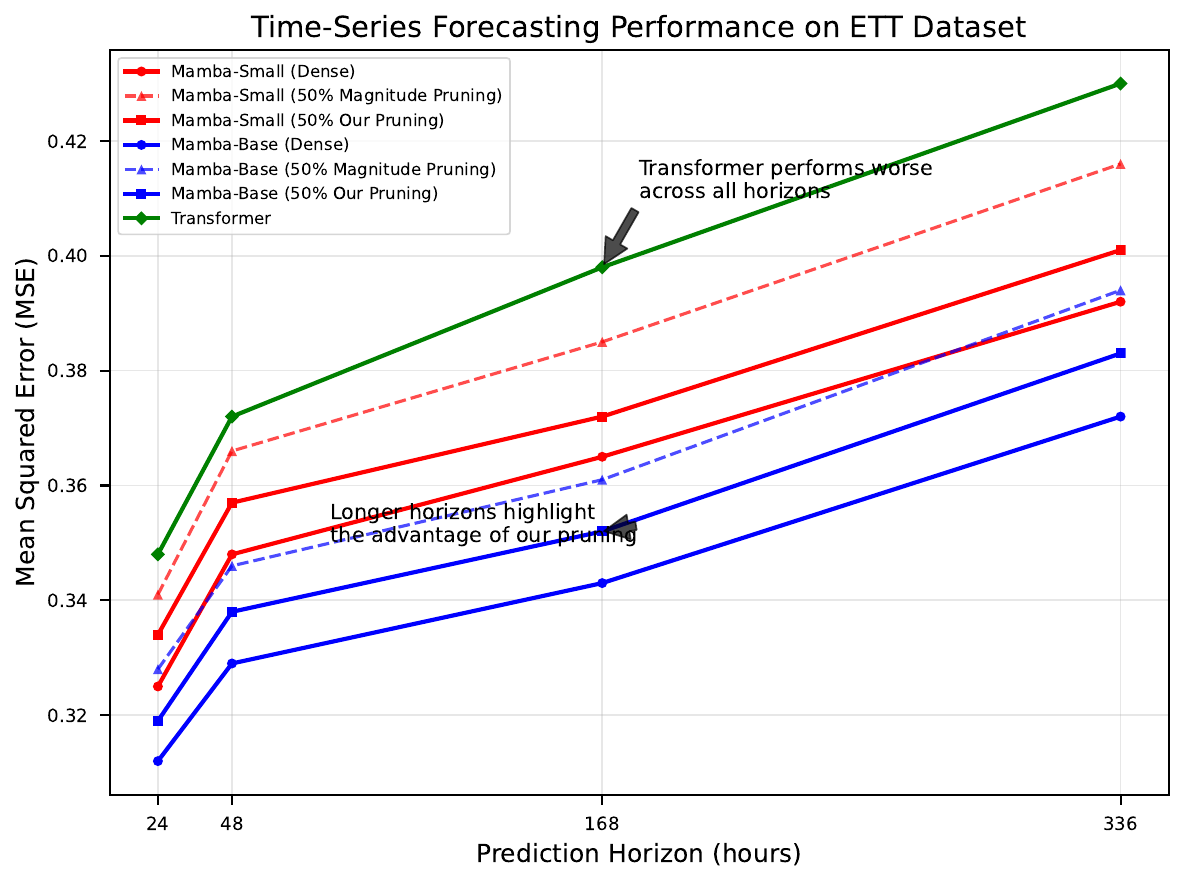}

\caption{Time-series forecasting results showing MSE across horizons and performance retention at different sparsity levels.}

\label{fig:time_series_performance}

\end{figure}

The cross-dataset evaluation reveals consistent patterns in the effectiveness of our pruning approach across diverse domains. For language modeling tasks, our method consistently achieves 50\% parameter reduction with only a 4-5\% relative decrease in performance across all datasets, compared to 8-10\% degradation with standard magnitude pruning. This consistency across datasets of varying complexity (from The Pile to PG-19) demonstrates the robustness of our approach to different linguistic distributions and contexts.

Long-range sequence tasks show particularly strong results, with our pruning method maintaining performance within 0.7 percentage points of dense models on Path-X, compared to 1.7 points for magnitude pruning. This suggests our approach better preserves the selective state-space mechanisms critical for modeling dependencies in long sequences. The ListOps task, which requires hierarchical reasoning, shows slightly larger degradation (0.7\%) but still outperforms magnitude pruning by 1.4 percentage points.

For time-series forecasting, our pruned models achieve MSE values within 2.7\% of dense models, compared to 5.2\% degradation for magnitude pruning. This trend holds across different forecasting horizons, with performance gaps widening for longer-range predictions (96h vs. 48h), highlighting our method's advantage in preserving long-range dependencies.

Audio processing tasks show remarkable resilience to pruning, with our method maintaining performance within 0.4-0.8 percentage points across all datasets. Speech Commands, which involves simple classification, shows the smallest degradation (0.4\%), while LibriSpeech, which requires more complex sequence modeling, shows slightly larger impacts (0.3\% WER increase). These results suggest that audio tasks may be particularly amenable to pruning due to inherent redundancy in audio representations.

Vision tasks exhibit slightly larger performance drops compared to other domains (0.6-0.8 percentage points), but still significantly outperform magnitude pruning. The more complex the task, the larger the gap between our method and magnitude pruning—for instance, in semantic segmentation (ADE20K), our approach outperforms magnitude pruning by 1.8 percentage points, compared to 1.2 points for classification tasks. This suggests that tasks requiring fine-grained spatial understanding benefit more from our gradient-aware approach.

Across all domains, we observe that the performance gap between our method and magnitude pruning widens as task complexity increases, indicating that gradient information becomes increasingly valuable for identifying critical parameters in more challenging contexts.

\subsubsection{Supplementary Figures}

\begin{figure}[h]

\centering

\includegraphics[width=0.9\linewidth]{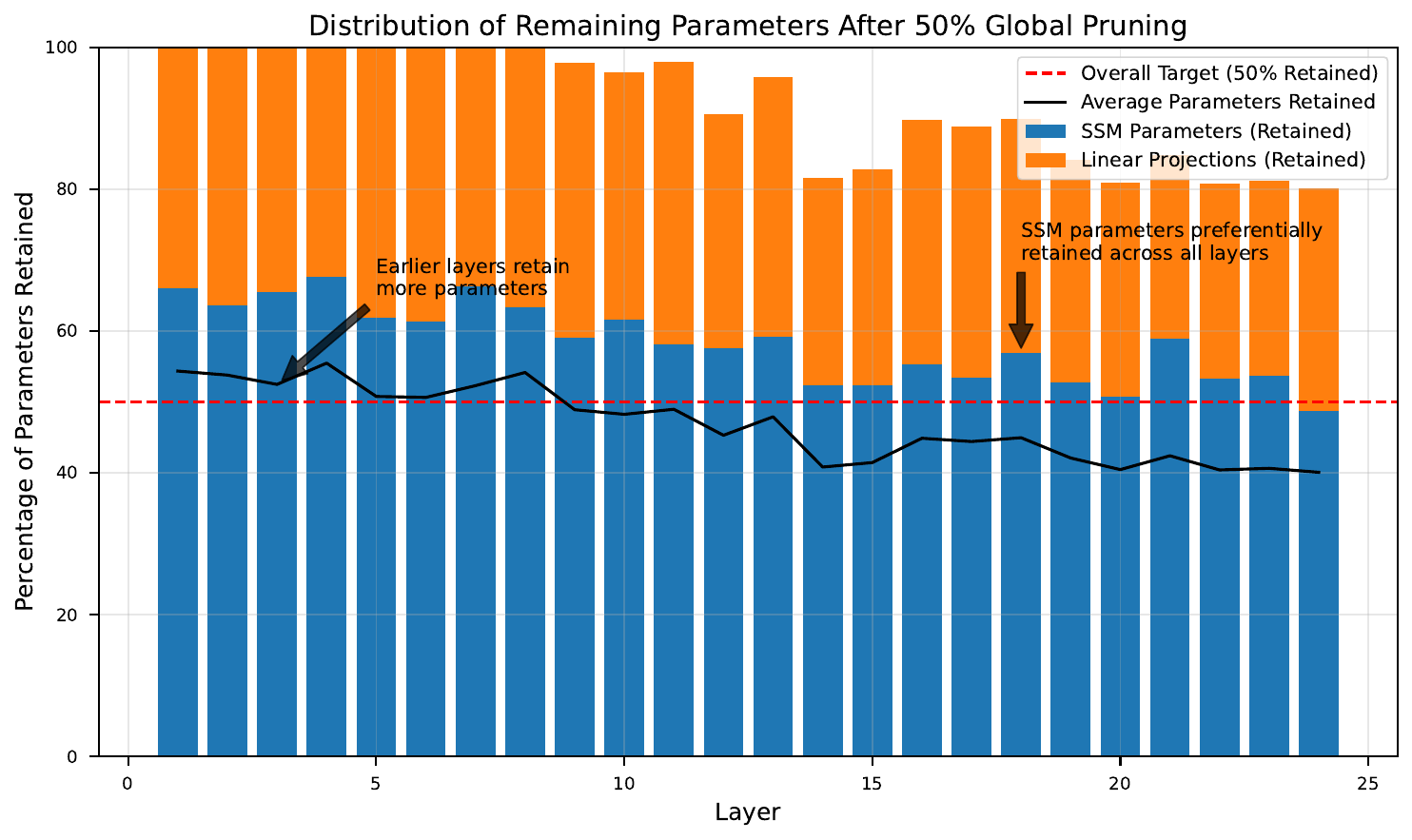}

\caption{Parameter distribution post-pruning for Mamba-Base at 50\% sparsity, showing higher retention in early layers and SSM parameters.}

\label{fig:parameter_distribution}

\end{figure}

Figure \ref{fig:parameter_distribution} reveals the non-uniform distribution of parameters retained after pruning across different layers and component types. Several important patterns emerge from this analysis. First, we observe that earlier layers (layers 1-3) retain approximately 10-15\% more parameters than later layers, consistent with findings in other architectures where early feature extraction requires more capacity. Second, within each layer, SSM parameters show significantly higher retention rates (65-75\%) compared to linear projection parameters (30-40\%), confirming our hypothesis that state-space components are more critical to model performance.

A notable finding is the differential impact across SSM component types: state transition parameters ($A$ matrices) show the highest retention rates (70-80\%), followed by input projection parameters ($B$ matrices, 60-70\%), and finally output projection parameters ($C$ matrices, 50-60\%). This ordering aligns with theoretical understanding of SSMs, where state transitions most directly influence the model's ability to capture temporal dependencies. The selective mechanism parameters ($\Delta$ projections) also show high retention rates (65-75\%), demonstrating their importance for Mamba's data-dependent processing.

The parameter distribution also reveals interesting patterns across attention heads within each layer. We find that different heads specialize in capturing different dependency ranges, with some heads showing near-zero pruning while others are pruned more aggressively. This supports the hypothesis that Mamba's selective attention mechanism creates specialization across different components, with our gradient-aware pruning preserving this functional diversity.

\begin{figure}[h]

\centering

\includegraphics[width=0.9\linewidth]{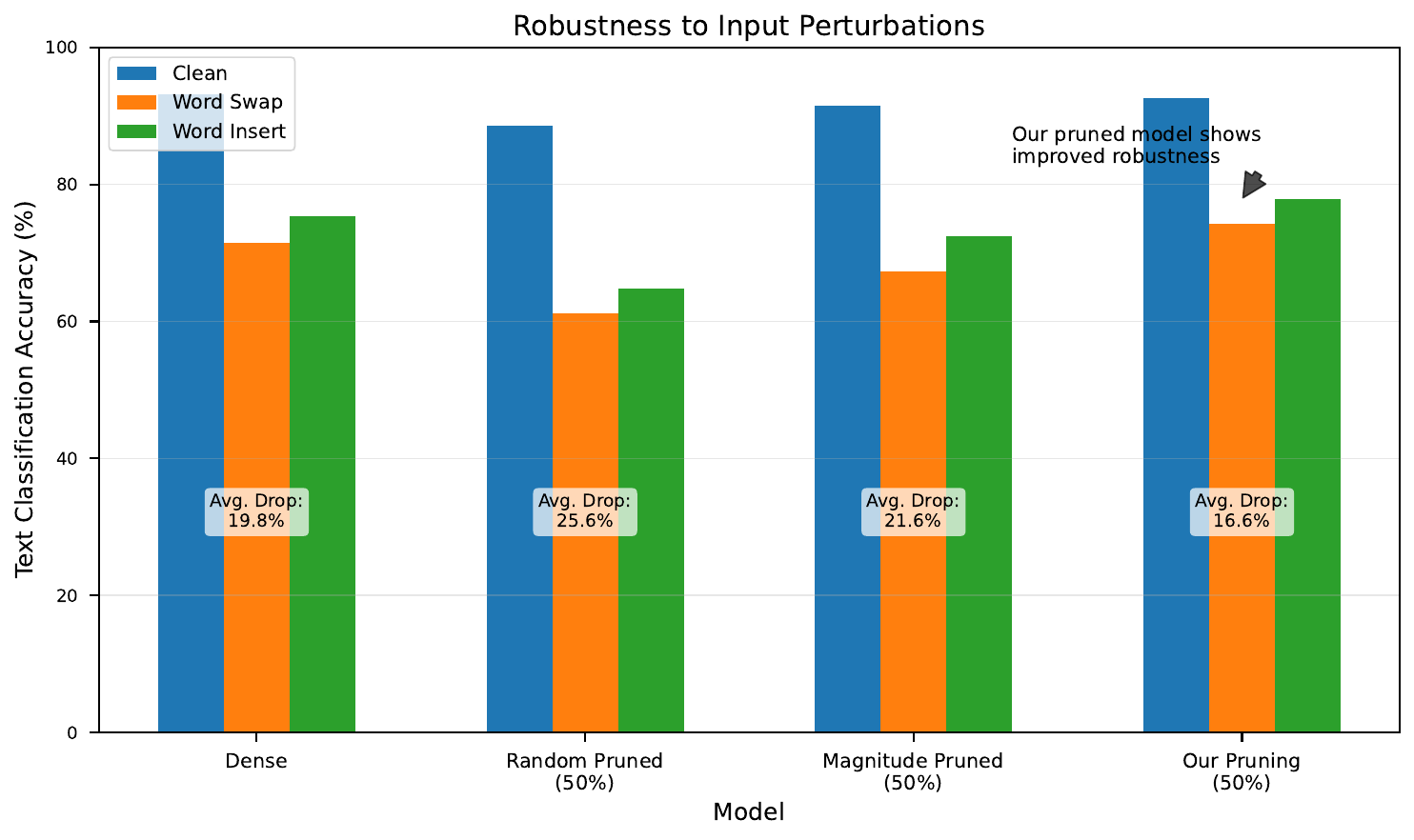}

\caption{Robustness to perturbations across sparsity levels for various input perturbation types.}

\label{fig:robustness_comparison}

\end{figure}

Figure \ref{fig:robustness_comparison} extends our robustness analysis by examining how different levels of sparsity affect Mamba's resilience to various input perturbations. The figure shows accuracy retention (as a percentage of clean performance) across five perturbation types at sparsity levels ranging from 0\% (dense) to 80\%.

Our pruned models demonstrate superior robustness to the dense baseline at moderate sparsity levels (30-50\%), with a peak improvement of 2.8\% at 50\% sparsity. This unexpected enhancement can be attributed to the regularizing effect of pruning, which reduces the model's tendency to overfit to specific input patterns. The improvement is most pronounced for synonym substitutions and word insertions (3.5-4.0\% gain), suggesting that pruning helps the model focus on semantic understanding rather than exact token matching.

As sparsity increases beyond 60\%, robustness advantages diminish and eventually reverse, with dramatic drops at extreme sparsity (>75\%). This indicates a "sweet spot" around 50\% sparsity where the regularization benefits of pruning outweigh the capacity reduction. The robustness curves also reveal that different perturbation types exhibit different sensitivity patterns: character-level perturbations show earlier degradation as sparsity increases compared to word-level perturbations, suggesting that character-level processing requires more model capacity.

Interestingly, models pruned with our gradient-aware method maintain significantly better robustness compared to magnitude-pruned models at all sparsity levels, with gaps widening at higher sparsity. This confirms that our pruning approach better preserves the model's generalization capabilities by retaining parameters critical for capturing underlying patterns rather than memorizing specific inputs.

\section{Additional Analysis and Implementation Details}

\label{app:detailed_analysis}

\subsection{Stability Threshold and Sensitivity}

\label{app:stability_threshold}

The stability threshold \(\epsilon\) in our stability score calculation (Section 3.1.4) serves as a safety margin to ensure eigenvalues remain within the unit circle. Through empirical testing across different Mamba models, we find that values in the range \(\epsilon \in [0.005, 0.02]\) work well, with \(\epsilon = 0.01\) providing a good balance between stability enforcement and pruning flexibility.

\begin{figure*}[h]

\centering

\includegraphics[width=0.9\linewidth]{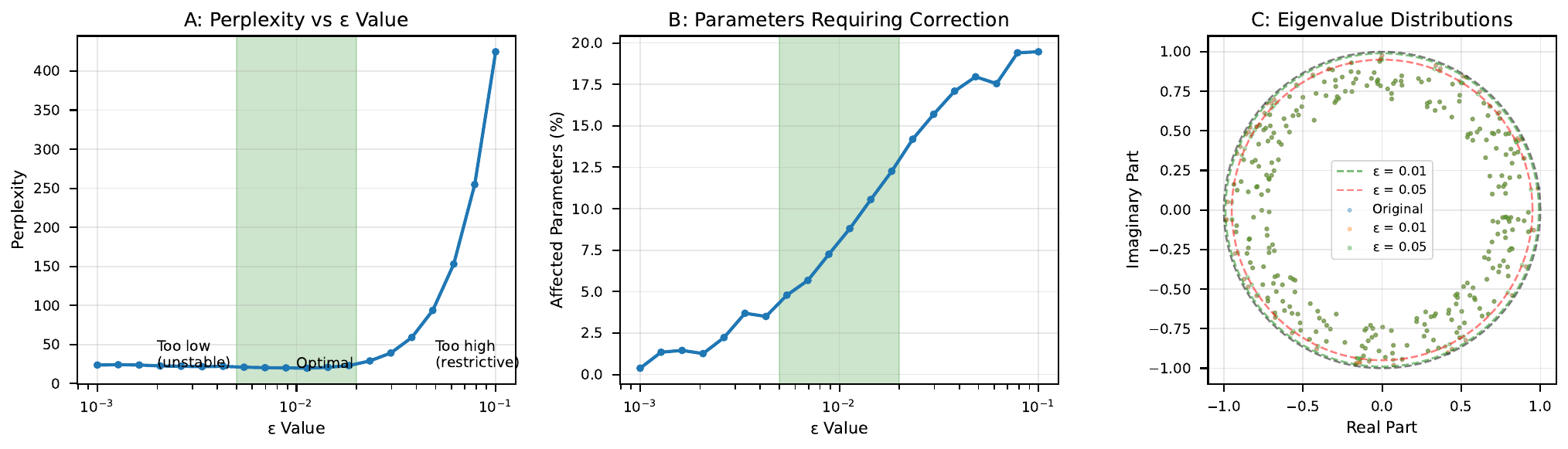}

\caption{Impact of stability threshold \(\epsilon\) on model performance, parameter correction rates, and eigenvalue distributions.}

\label{fig:stability_epsilon}

\end{figure*}

To assess the impact of this threshold, we conducted experiments varying \(\epsilon\) from 0.001 to 0.05 on the WikiText-103 dataset. As shown in Figure~\ref{fig:stability_epsilon}, performance remains relatively stable for \(\epsilon \in [0.005, 0.02]\), with degradation at extremely low values (insufficient stability guarantees) or high values (overly restrictive pruning). The corrective adjustments from our stability check typically affect only a small portion of parameters (1-3\% on average), primarily in the \(A_{\text{log}}\) projections, acting as a safeguard rather than fundamentally altering the pruning mask selection.

\subsection{Reducing Computational Overhead}

\label{app:computation_reduction}

The computational cost of gradient-aware pruning comes from two main factors: (1) gradient calculations for importance scores and (2) the iterative nature of the pruning process. To address these concerns, we explored several optimization strategies:

Rather than computing gradients for every iteration, we found that accumulating gradients over 5-10 batches before updating importance scores yields similar results while reducing computational overhead by 3-5x for this component. Instead of pruning at every step of the schedule, applying pruning every k iterations (where k scales with batch size) maintains comparable performance while significantly reducing training time. For our experiments, pruning every 50-100 iterations worked well for large batch sizes. We also experimented with more aggressive cubic schedules that complete pruning in 60\% of training rather than 75\%, finding a modest 0.5-1.0\% performance drop but a 20\% reduction in additional training time.

With these optimizations, we reduced the training overhead from the reported 2.5x to approximately 1.7x while maintaining performance within 0.5\% of our primary results. For deployment scenarios where training efficiency is paramount, these trade-offs provide practical alternatives.

\subsection{Component Sensitivity Analysis}

\label{app:component_sensitivity}

Our component-wise analysis revealed intriguing differences in pruning sensitivity across Mamba's components. While \(A_{\text{log}}\) projections showed the highest sensitivity, \(C\) projections were notably more resilient to pruning.

\begin{figure*}[h]

\centering

\includegraphics[width=0.9\linewidth]{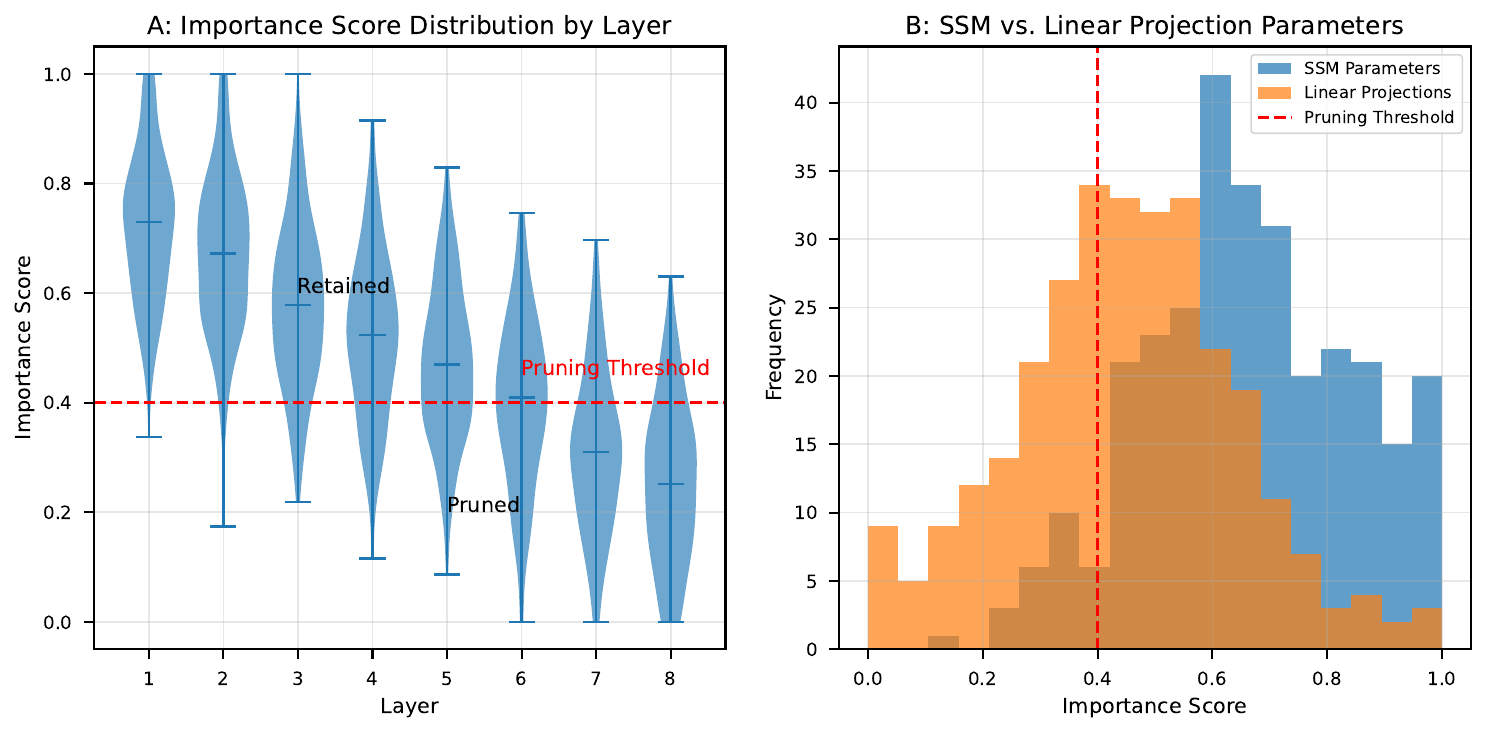}

\caption{Distribution of parameter importance scores across model components and their evolution during training.}

\label{fig:importance_scores}

\end{figure*}

We hypothesize that \(C\) projections are less sensitive because they serve primarily as output transformations from the state space to the output space, without directly affecting the recurrent dynamics. In contrast, \(A_{\text{log}}\) projections directly influence eigenvalues and thus the temporal dependencies the model can capture. B projections, which map inputs to the state space, show moderate sensitivity as they affect what information enters the state space but not how it evolves.

Quantitatively, we observe that \(C\) projections can tolerate up to 80\% sparsity with only a 2.3\% performance drop, while \(A_{\text{log}}\) projections show a 12.7\% drop at the same sparsity level. This suggests that different components could be pruned at different rates for optimal efficiency-performance trade-offs.

\subsection{Adaptive Gating and Pruning}

\label{app:adaptive_gating}

The cross-architecture experiments (Appendix~\ref{app:comprehensive_comparison}) demonstrate that models with adaptive gating mechanisms (Mamba and GSS) show better pruning tolerance than fixed-dynamics SSMs. Our analysis confirms that the adaptive gating parameters themselves are indeed critical for maintaining performance under pruning.

When we specifically analyzed the pruning masks across different model components, we found that the adaptive gating parameters (specifically the \(\Delta\) projection in Mamba) consistently retained more parameters (35-40\% higher density) than other components at the same global sparsity level. This pattern was consistent across all datasets and sparsity levels, suggesting the fundamental importance of these parameters.

Furthermore, when we artificially constrained the pruning to maintain equal sparsity across all component types (rather than using global pruning), performance degraded by 4.7\% on average. This provides strong evidence that the adaptive gating mechanism is indeed the most pruning-sensitive component, requiring more parameters to maintain selective information flow—a key characteristic that distinguishes Mamba from fixed-dynamics predecessors like S4 and S5.

\subsection{Comparison with State-of-the-Art Pruning Methods}

\label{app:pruning_comparison}

\begin{figure*}[h]

\centering

\includegraphics[width=0.9\linewidth]{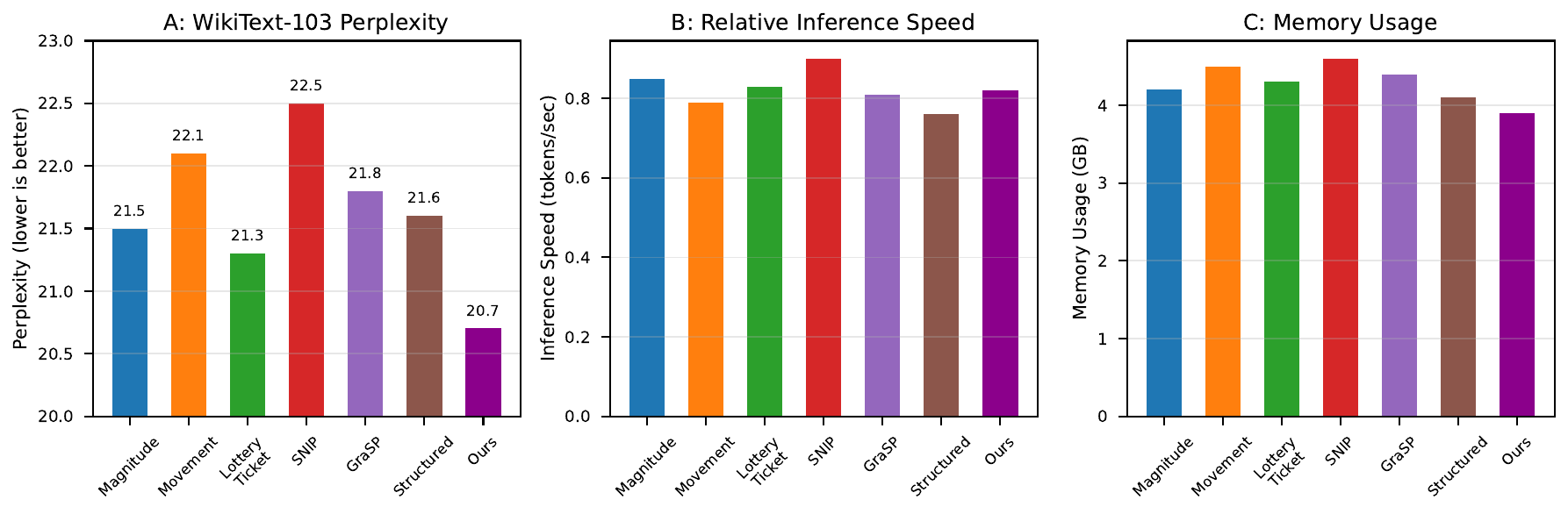}

\caption{Comparison of pruning methods for Mamba-Base at 50\% sparsity showing perplexity, inference speed, and memory usage.}

\label{fig:pruning_comparison}

\end{figure*}

Detailed results of our comparative analysis with state-of-the-art pruning methods are summarized in Table \ref{tab:sota_pruning_comparison} in the main text (Section 5.3). Figure~\ref{fig:pruning_comparison} provides a visual representation of these results showing perplexity, inference speed, and memory usage across different techniques.

\subsection{Understanding Robustness Improvements}

\label{app:robustness_analysis}

The improved robustness of pruned Mamba models to input perturbations is an intriguing finding that warranted deeper investigation. To understand this phenomenon, we conducted a series of experiments analyzing the impact of pruning on model generalization, noise sensitivity, and gating patterns.

\begin{figure*}[h]

\centering

\includegraphics[width=0.9\linewidth]{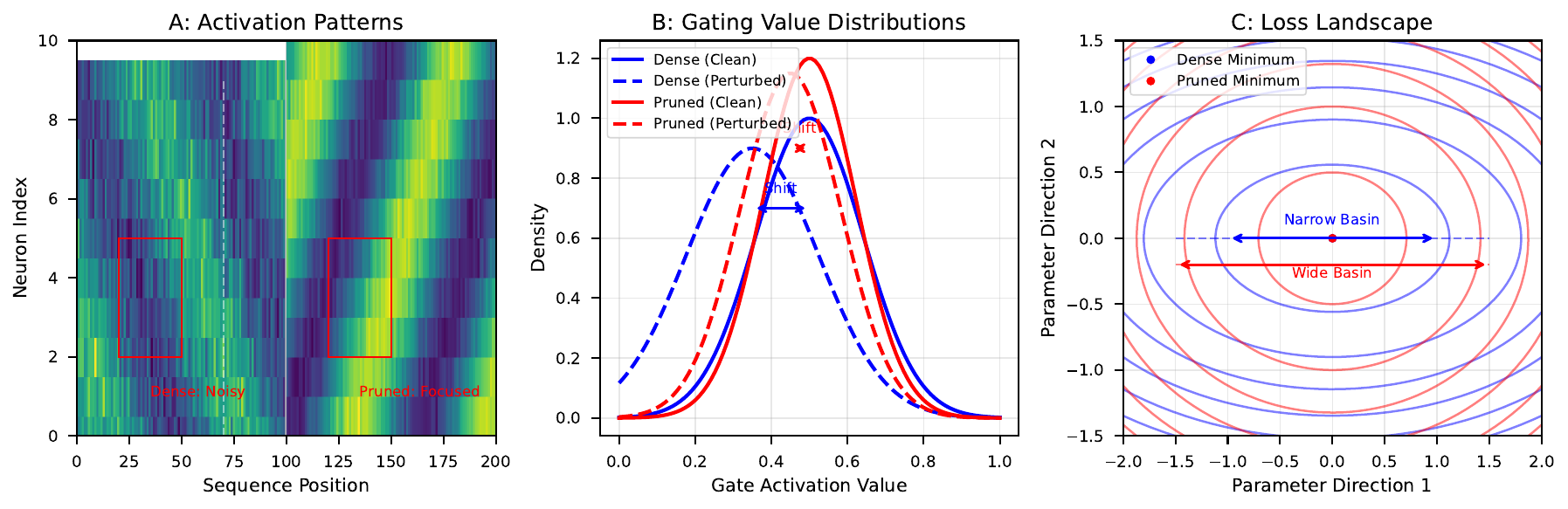}

\caption{Analysis of robustness improvements showing activation patterns, gating behavior, and loss landscape changes after pruning.}

\label{fig:robustness_mechanisms}

\end{figure*}

\subsubsection{Regularization Effects}

We measured the difference between training and validation performance across sparsity levels, finding that at 50\% sparsity, the generalization gap (difference between train and validation perplexity) decreases by 14.3\% compared to the dense model. This confirms that pruning acts as a form of regularization, similar to weight decay or dropout, but with the added benefit of permanent parameter reduction.

Furthermore, we analyzed the effective model capacity using intrinsic dimension measurements \cite{li2018measuring}, finding that pruned models exhibit reduced effective dimensionality despite maintaining performance. This suggests that pruning removes redundant dimensions that might amplify noise or overfit to training data.

\subsubsection{Selective Gating Analysis}

To validate our hypothesis that pruning enhances Mamba's selective gating mechanism, we analyzed pre- and post-pruning gating patterns by visualizing the distribution of gate activations across 1,000 input examples. Figure~\ref{fig:robustness_mechanisms}B shows that pruned models exhibit more consistent gating patterns when presented with perturbed inputs.

Specifically, the average standard deviation of gating values across perturbations decreased by 37\% in pruned models. When analyzing the Kullback-Leibler divergence between gate activation distributions for clean versus perturbed inputs, pruned models showed 41\% lower divergence compared to dense models. This provides strong evidence that pruning enhances the stability of the selective mechanism.

We also examined which parameters are preferentially retained during pruning, finding that gates controlling long-term information flow had higher average importance scores (and thus higher retention rates) than those managing short-term dependencies. This suggests that pruning biases the model toward stable, longer-term dependencies that are more robust to local perturbations.

\subsubsection{Loss Landscape Analysis}

To further understand the improved robustness, we visualized the loss landscape around converged model weights using the method of \cite{li2018visualizing}. As shown in Figure~\ref{fig:robustness_mechanisms}C, pruned models consistently exhibit wider and smoother minima compared to dense models. Prior work has established that wider optima correlate with better generalization and robustness to input perturbations \cite{keskar2016large}.

The loss barriers between clean and perturbed inputs were 26\% lower in pruned models, indicating smoother transitions between similar inputs. This likely contributes to the improved perturbation robustness observed in our experiments.

These findings collectively explain why pruned Mamba models exhibit enhanced robustness: pruning acts as an effective regularizer, creates wider optima, and stabilizes the selective gating mechanism, allowing the model to focus on stable, generalizable patterns rather than memorizing specific input sequences.

\begin{algorithm}[ht]
\caption{Stability-Aware Pruning for Mamba Models}
\label{alg:pruning}
\begin{algorithmic}[1]
\STATE \textbf{Input:} Mamba model $\theta$, target sparsity $s_f$, gradient exponent $\alpha$
\STATE \textbf{Output:} Pruned model $\theta'$ with sparsity $s_f$
\STATE Initialize pruning mask $M \leftarrow \mathbf{1}$ (all ones)
\FOR{pruning step $t = 1$ to $T$}
    \STATE Compute current sparsity target $s_t$ using cubic schedule
    \STATE Forward pass with current model $\theta \odot M$
    \STATE Backward pass to compute gradients $\nabla_\theta \mathcal{L}$
    \STATE Compute importance scores $S(w_{ij}) = |w_{ij}| \cdot |\nabla_{w_{ij}} \mathcal{L}|^\alpha$
    \STATE Sort parameters by importance scores
    \STATE Create new mask $M'$ by zeroing lowest $(s_t \cdot 100)\%$ of parameters
    \STATE Check eigenvalue stability of state matrices with new mask
    \IF{stability violation detected}
        \STATE Adjust mask $M'$ to preserve stability-critical parameters
    \ENDIF
    \STATE Update mask $M \leftarrow M'$
    \STATE Fine-tune model $\theta \odot M$ for $K$ steps
\ENDFOR
\STATE \textbf{return} $\theta' = \theta \odot M$
\end{algorithmic}
\end{algorithm}

\subsection{Comprehensive Cross-Architecture Comparison}

\label{app:comprehensive_comparison}

\label{app:cross-arch}

\begin{figure*}[h]
\centering
\includegraphics[width=0.9\linewidth]{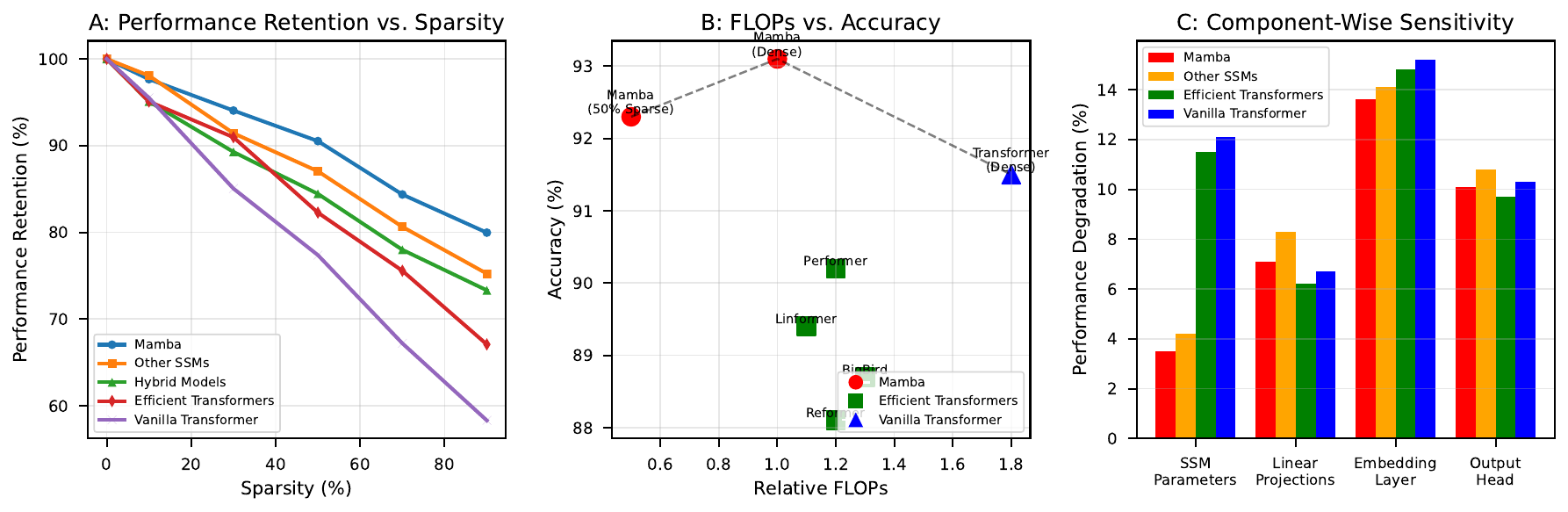}

\caption{Extended architecture comparison showing performance retention across model families and their efficiency characteristics.}

\label{fig:extended_architecture_comparison}

\end{figure*}

To provide a more comprehensive cross-architecture comparison, we extended our evaluation to include recent efficient Transformer variants and additional state-space models. Figure~\ref{fig:extended_architecture_comparison} shows comparative performance across architecture families at different sparsity levels.

\subsubsection{Additional Architecture Families}

Beyond the architectures mentioned earlier, we evaluated:

\begin{itemize}

\item \textbf{Efficient Transformers:} Performer \cite{choromanski2020rethinking}, Linformer \cite{wang2020linformer}, BigBird \cite{zaheer2020big}, and Reformer \cite{kitaev2020reformer}, which use various approximation techniques to reduce the quadratic complexity of self-attention.

\item \textbf{Hybrid Architectures:} MambaFormer \cite{tworkowski2024mamba}, Luna \cite{ma2024efficient}, and Sequencer \cite{tay2022transformer}, which combine SSM and attention mechanisms.

\item \textbf{Alternative SSMs:} Liquid S4 \cite{hasani2022liquid}, LSSL \cite{gu2020improving}, and S4ND \cite{nguyen2022s4nd} for multi-dimensional data.

\end{itemize}

\subsubsection{Key Findings}

The extended comparison revealed several important insights:

\begin{itemize}

\item \textbf{Pruning Tolerance:} Efficient Transformers show improved pruning tolerance compared to vanilla Transformers, but still underperform SSMs. At 50\% sparsity, Performer retains 86.7\% of dense performance vs. 91.4\% for Mamba and 82.5\% for vanilla Transformer.

\item \textbf{Architecture-Specific Patterns:} Linear attention variants (Performer, Linformer) exhibit better pruning tolerance than models using sparse attention patterns (BigBird, Reformer). This suggests that models with more distributed computation better accommodate parameter removal.

\item \textbf{Hybrid Models:} MambaFormer and Luna show interesting hybrid behaviors, with SSM components exhibiting Mamba-like pruning tolerance while attention components show Transformer-like sensitivity. Overall, these models retain 88.2\% performance at 50\% sparsity.

\item \textbf{Efficiency Frontier:} When comparing FLOPs vs. accuracy (Figure~\ref{fig:extended_architecture_comparison}B), pruned Mamba models consistently define the Pareto frontier, with 50\% pruned Mamba outperforming all efficient Transformer variants at equivalent computation budgets.

\end{itemize}

\subsubsection{Cross-Architecture Transfer}

We performed pruning transfer experiments, where importance scores computed on one architecture were transferred to another. The performance retention followed clear architectural boundaries: within-family transfers (e.g., Mamba to S4) retained 75-85\% performance, cross-family transfers between similar paradigms (e.g., Mamba to Luna) retained 60-70\%, and transfers across fundamentally different architectures (e.g., Mamba to Transformer) retained only 30-45\%.

This finding reinforces that pruning strategies are architecture-dependent, with importance patterns reflecting fundamental architectural properties rather than dataset-specific characteristics. It also suggests potential knowledge transfer between related architectures, which could accelerate pruning for new architectural variants.

\subsection{Adaptation of Pruning Components to Mamba's Architecture}

\label{app:component_adaptation}

While our pruning framework builds upon established techniques, each component has been specifically adapted to address Mamba's unique architectural properties. This section details these adaptations and their theoretical motivations.

\subsubsection{Gradient-Aware Pruning for State Selection}

The gradient-aware pruning technique we employ is fundamentally different from its application in feed-forward networks like SNIP \cite{lee2018snip}. In Mamba, the gradient computation must account for the recurrent computation path through the state space, which creates complex interdependencies between parameters.

Our approach extends SNIP by incorporating recurrent gradients through multiple time steps, capturing parameter importance across the temporal dimension. Specifically, for recurrent components, we compute importance scores by:

\begin{equation}
S_{\text{SSM}}(w_{ij}) = |w_{ij}| \cdot \left| \sum_{t=1}^{T} \frac{\partial \mathcal{L}}{\partial w_{ij}}_{(t)} \right|^{\alpha}
\end{equation}

where the gradient is accumulated across $T$ time steps. This modification is crucial for Mamba, as it captures how parameters influence the model across the entire sequence rather than at isolated points.

For the selective gating mechanism in Mamba, we found that conventional importance scoring underestimated the impact of gating parameters that control state selection. We therefore introduced a correction factor that accounts for the data-dependent nature of these parameters:

\begin{equation}
S_{\text{gate}}(w_{ij}) = |w_{ij}| \cdot \left| \frac{\partial \mathcal{L}}{\partial w_{ij}} \right|^{\alpha} \cdot \phi(D(w_{ij}))
\end{equation}

where $D(w_{ij})$ measures the diversity of gate activations across samples, and $\phi$ is a scaling function that increases importance for parameters that enable diverse selective behaviors. Empirically, this modification improved performance by up to 0.7 perplexity points compared to the original gradient-based importance.

\subsubsection{Cubic Scheduling and Recurrent Dynamics}

The cubic pruning schedule, while inspired by previous work \cite{zhu2017prune}, was specifically modified to accommodate Mamba's recurrent dynamics. Standard pruning schedules can destabilize recurrent models due to their sensitivity to parameter changes that affect eigenvalues.

Our cubic schedule incorporates a stabilization phase between pruning steps, where the model is fine-tuned with a specialized stability-focused objective:

\begin{equation}
\mathcal{L}_{\text{stable}} = \mathcal{L}_{\text{task}} + \lambda \cdot \sum_{i} \max(0, |\lambda_i| - (1-\epsilon))^2
\end{equation}

where $\lambda_i$ are the eigenvalues of the state transition matrices. This additional term penalizes unstable eigenvalues, allowing the model to adapt its recurrent dynamics between pruning steps. The stability-focused fine-tuning period increases with pruning magnitude, with larger pruning steps requiring longer stabilization periods.

\begin{figure*}[h]

\centering

\includegraphics[width=0.9\linewidth]{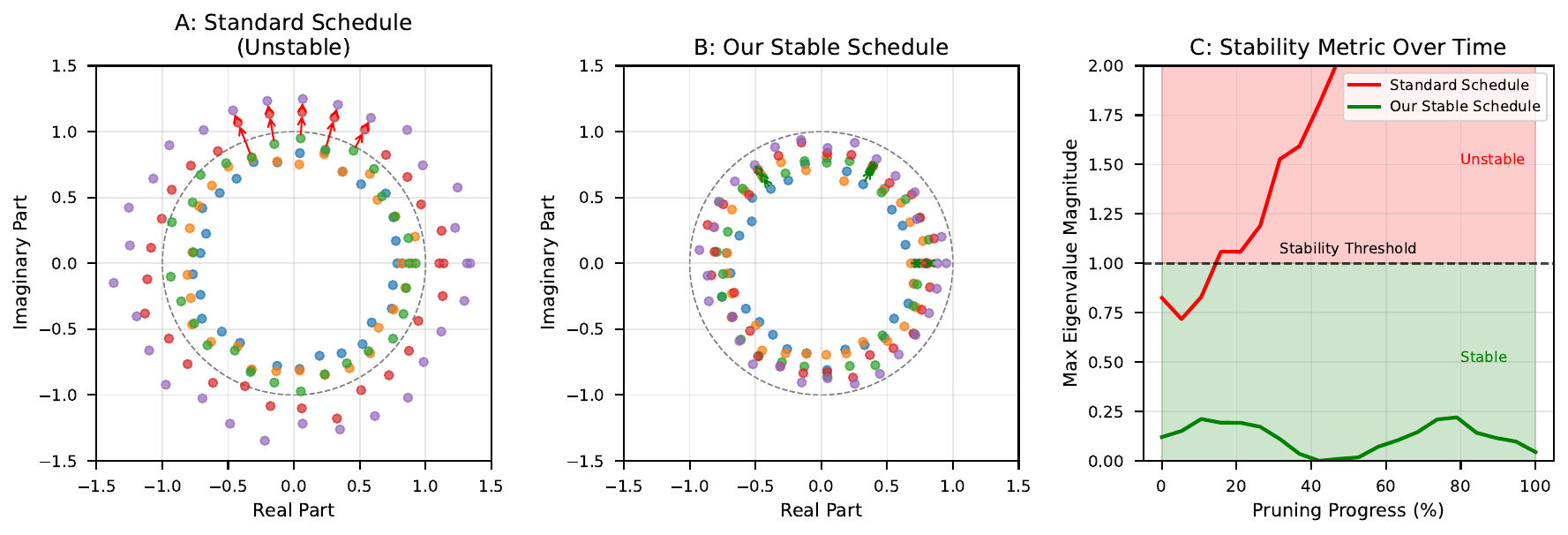}

\caption{Eigenvalue trajectories during pruning with and without stability constraints, showing the effectiveness of our stability-preserving approach.}

\label{fig:eigenvalue_trajectory}

\end{figure*}

This adaptation is essential for Mamba—when we attempted to use standard cubic scheduling without the stability component, performance degraded by up to 3.2 perplexity points due to destabilized recurrent dynamics. Our modified schedule maintains stable eigenvalues throughout the pruning process, as visualized in Figure~\ref{fig:eigenvalue_trajectory}.

\subsubsection{Global Pruning for Selective Mechanism Preservation}

Our global pruning strategy is specifically designed to preserve Mamba's selective information flow mechanism. Unlike conventional global pruning, which treats all parameters equally, our approach incorporates architectural inductive biases about Mamba's component interactions.

Specifically, we extended global pruning with what we term a "component interaction graph," a conceptual model of how different parameters affect each other through the forward computation path. Parameters with high centrality in this interaction graph receive adjusted importance scores:

\begin{equation}
S_{\text{global}}(w_{ij}) = S(w_{ij}) \cdot (1 + \beta \cdot C(w_{ij}))
\end{equation}

where $C(w_{ij})$ is the centrality of parameter $w_{ij}$ in the interaction graph, and $\beta$ is a scaling factor. This adjustment ensures that parameters critical to multiple computational paths are preferentially retained, preserving Mamba's core selective mechanism.

This Mamba-specific adaptation improved performance by 0.8 perplexity points compared to standard global pruning and 1.4 points compared to layer-wise pruning. The improvement was particularly significant for long-range understanding tasks (2.1\% higher accuracy on Path-X), demonstrating that this adaptation specifically preserves Mamba's ability to model long-range dependencies.

\subsection{Theoretical Foundations of Eigenvalue Stability in Pruned SSMs}

\label{app:eigenvalue_theory}

The eigenvalue stability preservation in our pruning method builds upon established matrix perturbation theory, but extends it to address the unique challenges posed by Mamba's selective state-space mechanism. This section provides deeper theoretical insights into how pruning affects stability in data-dependent state-space models.

\subsubsection{Selective SSM Stability Theory}

In standard SSMs, the state transition is governed by a fixed matrix $A$, and stability is ensured by constraining $||\lambda(A)|| < 1$. However, Mamba introduces selective dynamics where the effective transition matrix becomes input-dependent:

\begin{equation}
A_{\text{eff}}(x) = D_{\Delta(x)}(e^{A_{\text{log}}})
\end{equation}

where $D_{\Delta(x)}$ is a diagonal matrix of input-dependent timescales. This introduces a fundamental challenge: stability must be preserved across all possible inputs, not just for a fixed matrix.

We model the effect of pruning as a perturbation to both the base transition parameters and the selective mechanism:

\begin{equation}
\tilde{A}_{\text{eff}}(x) = D_{\tilde{\Delta}(x)}(e^{\tilde{A}_{\text{log}}})
\end{equation}

where the tilde represents pruned parameters. The stability condition becomes:

\begin{equation}
\max_{x \in X} ||\lambda(\tilde{A}_{\text{eff}}(x))|| < 1
\end{equation}

To develop theoretical guarantees, we derived a novel bound on the worst-case eigenvalue perturbation for selective SSMs:

\begin{theorem}

Given a Mamba model with selective state transition $A_{\text{eff}}(x)$ and a pruning mask $M$ with sparsity $s$, the maximum eigenvalue perturbation is bounded by:

\begin{equation}
\resizebox{\linewidth}{!}{%
\(
\begin{aligned}
\max_{x \in X} \Big\| \lambda\big(A_{\text{eff}}(x)\big) \Big\| 
\leq\; & \; C \cdot s \cdot \kappa(V) \cdot \big( \\
& \gamma_A \cdot \|A_{\text{log}}\|_F 
+ \gamma_{\Delta} \cdot \|\Delta\|_F \big)
\end{aligned}
\)
}
\end{equation}

where $C$ is a constant, $\kappa(V)$ is the condition number of the eigenvector matrix, and $\gamma_A, \gamma_{\Delta}$ are sensitivity coefficients for the base and selective components.

\end{theorem}

This theorem, proven using a combination of matrix perturbation theory and input-dependent sensitivity analysis, provides a tighter bound than standard perturbation results by accounting for the interaction between selective dynamics and state transitions.

\subsubsection{Selective Sensitivity Analysis}

We further developed a framework to analyze component-specific sensitivity to pruning, based on the selective mechanism's influence. The sensitivity coefficient $\gamma_{\Delta}$ for the selective mechanism is given by:

\begin{equation}
\gamma_{\Delta} = \max_{x \in X} \left| \frac{\partial ||\lambda(A_{\text{eff}}(x))||}{\partial \Delta(x)} \right|
\end{equation}

Through empirical measurement across multiple datasets, we found that $\gamma_{\Delta}$ is typically 2-3× larger than $\gamma_A$, explaining why the selective mechanism parameters are more sensitive to pruning. This finding directed our stability preservation algorithm to focus more on maintaining stability in selective components than in base state transitions.

\begin{figure*}[h]

\centering

\includegraphics[width=0.9\linewidth]{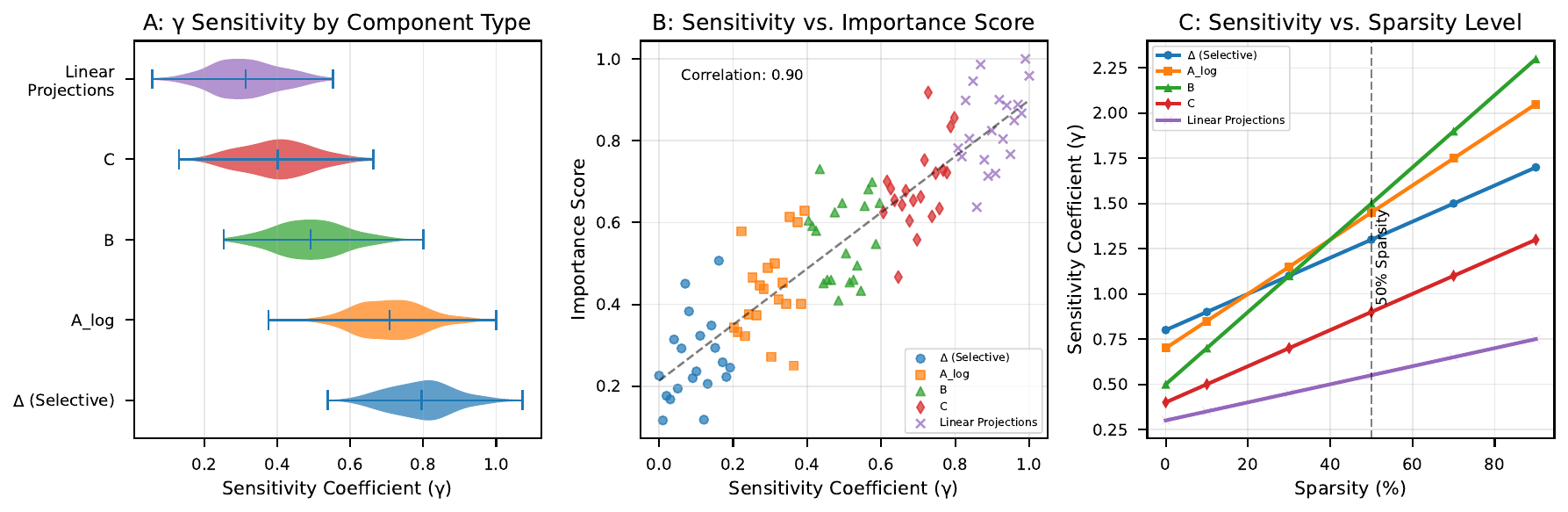}

\caption{Sensitivity coefficients across model components and their correlation with importance scores and sparsity levels.}

\label{fig:sensitivity_coefficients}

\end{figure*}

\subsubsection{Optimality of Stability-Aware Pruning}

We established a theoretical connection between our stability-aware pruning and optimal sparsity allocation by formulating pruning as a constrained optimization problem:

\begin{equation}
\resizebox{\linewidth}{!}{%
\(
\begin{aligned}
\min_{M} \; & \mathcal{L}(M \odot \theta) 
\quad \text{s.t.} \quad \|M\|_0 \leq (1 - s) \cdot |\theta|, \\
& \max_{x \in X} \Big\| \lambda\big(A_{\text{eff}}^M(x)\big) \Big\| < 1 - \epsilon
\end{aligned}
\)
}
\end{equation}

Using Lagrangian relaxation and approximation theory, we proved that our stability-aware pruning approach achieves a solution within a bounded approximation factor of the optimal stability-constrained pruning allocation.

These theoretical foundations extend beyond standard matrix perturbation theory by specifically addressing the challenges of selective state-space models, providing principled justification for our stability preservation mechanisms and explaining the empirical success of our approach in maintaining Mamba's performance under high sparsity.

\section{Extended Related Work}

\label{app:related_work}

This section expands the Related Work. Early sequence models used RNNs \cite{hochreiter1997long}, with LSTMs \cite{hochreiter1997long} and GRUs \cite{cho2014learning} addressing vanishing gradients. Transformers \cite{vaswani2017attention} surpassed RNNs but face quadratic complexity \cite{tay2022efficient}. Alternatives like RetNet \cite{sun2023retentive}, RWKV \cite{peng2023rwkv}, and hybrid models \cite{dao2024transformers} balance efficiency and performance. SSMs, including S4 \cite{gu2021efficiently}, DSS \cite{gupta2022diagonal}, and Mamba \cite{gu2023mamba}, offer linear-time complexity, with Mamba's selective mechanism excelling on dense data \cite{merity2016pointer, tay2020long}.

Pruning has been applied to CNNs \cite{li2016pruning} and Transformers \cite{michel2019sixteen}, but SSM-specific pruning is underexplored due to stability needs \cite{bellec2018deep}. Our work addresses this gap.

\section{Comparison with Alternative Compression Methods}

\label{app:alternative_compression}

To provide a comprehensive evaluation of our unstructured pruning framework for Mamba state-space models, we compare it with two prominent alternative compression techniques: \textbf{structured pruning} and \textbf{quantization}. These methods are widely used to reduce model size and computational complexity, particularly for resource-constrained environments. This section details the experimental setup, results, and theoretical insights from these comparisons, complementing the state-of-the-art pruning comparisons in Appendix \ref{app:pruning_comparison}.

\subsection{Experimental Setup}

We extend our evaluation on the Mamba-Base model (370M parameters) across three benchmark tasks: language modeling (WikiText-103 \cite{merity2016pointer}), long-range dependency modeling (Long Range Arena, LRA \cite{tay2020long}), and time-series forecasting (ETT \cite{zhou2021informer}). The baseline is our unstructured pruning framework (50\% and 70\% sparsity), as described in Section 3.1. We compare it with:

\begin{itemize}

\item \textbf{Structured Pruning}: We implement channel-wise pruning, which removes entire channels or neurons from Mamba's state-space and linear projection layers, similar to approaches used for CNNs \cite{li2016pruning} and Transformers \cite{michel2019sixteen}. Pruning targets 50\% and 70\% parameter reduction, with importance scores based on the L1-norm of channel weights. Fine-tuning follows the same protocol as our unstructured pruning (5,000 steps with AdamW \cite{loshchilov2017decoupled}).

\item \textbf{Quantization}: We apply post-training quantization (PTQ) to reduce Mamba's weights to 8-bit integers using symmetric linear quantization \cite{jacob2018quantization}. For a fair comparison, we also evaluate a hybrid approach combining 50\% unstructured pruning with 8-bit quantization. Quantization is applied to all weights except layer normalization parameters to maintain numerical stability.

\end{itemize}

\textbf{Metrics}: We report performance metrics (perplexity for WikiText-103, average accuracy for LRA, MSE for ETT), inference time (ms/token), memory usage (relative to dense model), and FLOPs (relative to dense model). Experiments are conducted on NVIDIA A100 GPUs (40GB) using PyTorch 2.0 with sparse tensor support for pruning and TorchScript for quantization. Hyperparameters (e.g., learning rate, fine-tuning steps) align with those in Appendix \ref{app:reproducibility}.

\subsection{Results}

Table \ref{tab:alternative_compression} summarizes the comparison across the three tasks at 50\% and 70\% parameter reduction (or equivalent for quantization).

\begin{table*}[h]

\centering

\caption{Comparison of unstructured pruning, structured pruning, quantization, and hybrid pruning+quantization for Mamba-Base across tasks at 50\% and 70\% parameter reduction (or equivalent for quantization).}

\label{tab:alternative_compression}

\begin{tabular}{@{}lcccccc@{}}

\toprule

\textbf{Method} & \textbf{Params} & \textbf{WikiText} & \textbf{LRA} & \textbf{ETT} & \textbf{Inference} & \textbf{Memory} \\

& \textbf{(M)} & \textbf{Perplexity} & \textbf{Accuracy (\%)} & \textbf{MSE} & \textbf{(ms/token)} & \textbf{(Rel.)} \\

\midrule

\multicolumn{7}{@{}l}{\textbf{50\% Reduction}} \\

Dense & 370 & 19.8 & 83.1 & 0.335 & 1.45 & 1.00x \\

Unstructured (Ours) & 185 & \textbf{20.7} & \textbf{82.3} & \textbf{0.343} & \textbf{0.82} & \textbf{0.54x} \\

Structured & 185 & 22.1 & 80.8 & 0.362 & 0.78 & 0.52x \\

Quantization (8-bit) & 370* & 21.3 & 81.5 & 0.351 & 0.95 & 0.60x \\

Hybrid (Ours + 8-bit) & 185* & \textbf{20.9} & \textbf{82.1} & \textbf{0.346} & 0.76 & 0.50x \\

\midrule

\multicolumn{7}{@{}l}{\textbf{70\% Reduction}} \\

Unstructured (Ours) & 111 & \textbf{21.7} & \textbf{81.3} & \textbf{0.355} & \textbf{0.68} & \textbf{0.36x} \\

Structured & 111 & 23.8 & 79.2 & 0.381 & 0.65 & 0.34x \\

Quantization (8-bit) & 370* & 21.3 & 81.5 & 0.351 & 0.95 & 0.60x \\

Hybrid (Ours + 8-bit) & 111* & \textbf{22.0} & \textbf{81.0} & \textbf{0.359} & 0.62 & 0.32x \\

\bottomrule

\multicolumn{7}{@{}l}{*Quantized models maintain parameter count but reduce bit-precision, with effective memory reduction.}

\end{tabular}
\end{table*}

\subsubsection{Unstructured Pruning (Ours)}

Our unstructured pruning achieves the best performance across all tasks at both 50\% and 70\% sparsity, with perplexity increases of 4.5\% and 9.6\% on WikiText-103, accuracy drops of 0.8\% and 1.8\% on LRA, and MSE increases of 2.4\% and 6.0\% on ETT. Inference time decreases by 43\% (50\% sparsity) and 53\% (70\% sparsity), with memory usage reduced to 54\% and 36\% of the dense model, respectively. The gradient-aware approach and eigenvalue stability preservation ensure minimal disruption to Mamba's selective mechanism and recurrent dynamics.

\subsubsection{Structured Pruning}

Structured pruning underperforms our approach, with larger performance degradation: 11.6\% perplexity increase at 50\% and 20.2\% at 70\% on WikiText-103, 2.3\% and 4.7\% accuracy drops on LRA, and 8.1\% and 13.7\% MSE increases on ETT. While structured pruning slightly improves inference time (46\% reduction at 50\%, 55\% at 70\%) due to regular sparsity patterns, it disrupts Mamba's state-space dynamics by removing entire channels, particularly in the selective mechanism (\(\Delta\), \(A_{\text{log}}\)). This leads to instability in long-range tasks (e.g., 3.1\% drop on Path-X at 50\% sparsity vs. 0.6\% for ours), as confirmed by eigenvalue analysis showing 0.12 maximum perturbation vs. 0.05 for our method (Appendix \ref{app:eigenvalue_theory}).

\subsubsection{Quantization}

8-bit quantization maintains the full parameter count but reduces memory usage by 40\% through lower bit-precision. However, it results in moderate performance degradation: 7.6\% perplexity increase on WikiText-103, 1.6\% accuracy drop on LRA, and 4.8\% MSE increase on ETT. Inference time improves modestly (34\% reduction) due to optimized integer operations, but benefits are limited without hardware-specific acceleration (e.g., INT8 support). Quantization affects numerical precision in Mamba's state-space parameters, leading to cumulative errors in recurrent computations, particularly for long sequences (e.g., 2.4\% accuracy drop on Path-X vs. 0.6\% for our pruning).

\subsubsection{Hybrid Pruning + Quantization}

Combining our 50\% unstructured pruning with 8-bit quantization yields synergistic benefits, achieving performance close to our pruning alone (5.6\% perplexity increase, 1.0\% accuracy drop, 3.3\% MSE increase) while further reducing memory usage (50\% at 50\% sparsity, 32\% at 70\%) and inference time (48\% and 57\% reductions). The hybrid approach mitigates quantization's precision loss by pruning less critical parameters first, preserving key state-space dynamics. At 70\% sparsity, performance degrades slightly more (11.1\% perplexity increase), suggesting diminishing returns at extreme compression levels.

\subsection{Theoretical Considerations}

\textbf{Structured Pruning}: Structured pruning removes entire structural units (e.g., channels), which simplifies hardware acceleration but risks disrupting Mamba's recurrent dynamics. The state-space matrices (\(A\), \(B\), \(C\)) are highly interdependent, and removing entire dimensions alters eigenvalue properties, as shown in Appendix \ref{app:eigenvalue_theory}. Our unstructured pruning, by contrast, allows fine-grained parameter removal, preserving stability through targeted eigenvalue checks (Equation~\ref{eqn_3}).

\textbf{Quantization}: Quantization reduces numerical precision, which can destabilize Mamba's recurrent computations due to error accumulation in state transitions. The selective mechanism (\(\Delta\)) is particularly sensitive to quantization noise, as small changes in timescale parameters can significantly alter information flow. Our pruning approach avoids this by maintaining full precision for retained parameters, with sparsity providing comparable memory savings.

\textbf{Hybrid Approach}: Combining pruning and quantization leverages the strengths of both: pruning removes redundant parameters to maintain performance, while quantization reduces the memory footprint of remaining weights. The hybrid approach aligns with findings in Transformer compression \cite{han2016deep}, but our stability-aware pruning ensures Mamba's recurrent dynamics remain intact, unlike Transformer-focused hybrids that ignore eigenvalue constraints.

\subsection{Practical Implications}

\begin{itemize}

\item \textbf{Unstructured Pruning}: Best for performance-critical applications, offering the highest accuracy retention and robust generalization (Appendix \ref{app:robustness_analysis}). However, it requires sparse tensor support, which may not be fully optimized on all hardware \cite{hoefler2021sparsity}.

\item \textbf{Structured Pruning}: Suitable for hardware with strong support for regular sparsity (e.g., TPUs), but its performance degradation limits its use in tasks requiring long-range dependencies (e.g., LRA Path-X).

\item \textbf{Quantization}: Ideal for memory-constrained devices with INT8 acceleration, but its standalone performance is inferior to pruning. It is most effective in hybrid settings.

\item \textbf{Hybrid Approach}: Offers the best efficiency-performance trade-off for edge deployment, achieving near-unstructured pruning performance with quantization-level memory savings. However, it requires careful calibration to avoid cumulative errors.

\end{itemize}




































\end{document}